\documentclass[11pt]{article}

\usepackage[final]{acl}
\usepackage{subcaption}
\usepackage{times}
\usepackage{latexsym}
\usepackage{enumitem}
\usepackage[T1]{fontenc}
\usepackage[utf8]{inputenc}

\usepackage{microtype}
\usepackage{inconsolata}

\usepackage{graphicx}
\usepackage{booktabs}
\usepackage{amsmath}
\usepackage{amssymb}
\usepackage{float}
\usepackage{multirow}

\title{Better and Worse with Scale: How Contextual Entrainment Diverges with Model Size}

\author{
  \textbf{Dikshant Kukreja\textsuperscript{1}, Kshitij Sah\textsuperscript{1}, Gautam Gupta\textsuperscript{1}, Avinash Anand\textsuperscript{4},} \\
  \textbf{Rajiv Ratn Shah\textsuperscript{1}, Zhengkui Wang\textsuperscript{4}, Aik Beng Ng\textsuperscript{3}, Erik Cambria\textsuperscript{2}} \\
  \vspace{0.2cm} \\
  \textsuperscript{1}IIIT Delhi, India \qquad
  \textsuperscript{2}Nanyang Technological University \\
  \textsuperscript{3}NVIDIA \qquad
  \textsuperscript{4}Singapore Institute of Technology \\
}

\begin{document}
\maketitle
\begin{abstract}
Larger language models become simultaneously better and worse at handling contextual information---better at ignoring false claims, worse at ignoring irrelevant tokens. We formalize this apparent paradox through the first scaling laws for contextual entrainment, the tendency of models to favor tokens that appeared in context regardless of relevance. Analyzing the Cerebras-GPT (111M--13B) and Pythia (410M--12B) model families, we find entrainment follows predictable power-law scaling, but with opposite trends depending on context type: semantic contexts show decreasing entrainment with scale, while non-semantic contexts show increasing entrainment. Concretely, the largest models are four times more resistant to counterfactual misinformation than the smallest, yet simultaneously twice as prone to copying arbitrary tokens. These diverging trends, which replicate across model families, suggest that semantic filtering and mechanical copying are functionally distinct behaviors that scale in opposition---scaling alone does not resolve context sensitivity, it reshapes it.
\end{abstract}

\section{Introduction}

Large language models increasingly rely on retrieved or user-provided context for response generation; yet, this reliance introduces a fundamental vulnerability: models can be distracted by contextual information regardless of its relevance or accuracy. This problem manifests across retrieval-augmented generation systems, where noisy or adversarial passages degrade output quality \citep{gao2023retrieval, fang2024enhancing}, and in long-context settings, where irrelevant information degrades attention on relevant tokens \citep{liu2024lost}. Understanding how models process contextual information is therefore critical for deploying robust systems.

\citet{niu2025llama} formalized \textit{contextual entrainment}: the tendency of language models to favor tokens that appeared in context solely due to their presence, regardless of semantic relevance. They quantify this using logit shifts for distractor token $d$ and gold token $g$:
\[
\Delta_t = \text{logit}(t \mid \text{ctx}) - \text{logit}(t \mid \varnothing), \quad t \in \{d, g\}.
\]
A positive $\Delta_d$ indicates entrainment—the model boosts a token simply because it appeared in context. Consider the query \textit{``The capital of Germany is \underline{\hspace{0.5cm}}''} (gold $g$ = \textbf{Berlin}). Four context conditions each embed a distractor $d$:

\begin{itemize}[noitemsep, topsep=2pt, leftmargin=*]
    \item \textbf{Related}: \textit{``The Eiffel Tower is in Paris.''} ($d$ = Paris). $\Delta_d > 0$ reflects semantic association.
    \item \textbf{Irrelevant}: \textit{``The water is warm.''} ($d$ = warm). $\Delta_d > 0$ without semantic justification.
    \item \textbf{Random}: \textit{``Calculator.''} ($d$ = Calculator). $\Delta_d > 0$ reveals pure mechanistic copying.
    \item \textbf{Counterfactual}: \textit{``The capital of Germany is Munich.''} ($d$ = Munich). $\Delta_d > 0$ and $\Delta_g < 0$ indicate susceptibility to misinformation.
\end{itemize}
Experimentally, context is prepended to the query (e.g., \texttt{``Calculator. The capital of Germany is''}). Detailed prompt templates are provided in Appendix~\ref{app:examples}.

\noindent \citet{niu2025llama} find that models exhibit $\Delta_d > 0$ across all conditions, demonstrating that entrainment occurs regardless of semantic relevance. Crucially, the magnitude differs: Related and Counterfactual contexts, which carry semantic content about the query, produce stronger entrainment than Random and Irrelevant contexts, which are semantically mismatched, suggesting two distinct dynamics—a \textit{mechanistic} level where any previously-seen token receives elevated probability, consistent with induction head circuits \citep{olsson2022context}, and a \textit{semantic} level where context relevance modulates entrainment strength.

\begin{table*}[t]
\centering
\small

% Part (a): Distractor Entrainment
\begin{subtable}{0.49\textwidth}
\centering
\setlength{\tabcolsep}{4pt}
\begin{tabular}{@{}lcccc@{}}
\toprule
\textbf{Context} & $b$ & 95\% CI & $R^2$ & $p$ \\
\midrule
Counterfactual & $-$0.330 & [$-$0.44, $-$0.22] & 0.926 & 5e-04 \\
Related        & $-$0.135 & [$-$0.16, $-$0.11] & 0.977 & 3e-05 \\
Irrelevant     & $+$0.091 & [$+$0.05, $+$0.13] & 0.879 & 2e-03 \\
Random         & $+$0.217 & [$+$0.14, $+$0.30] & 0.905 & 1e-03 \\
\bottomrule
\end{tabular}
\caption{Distractor Entrainment ($\Delta_{\mathrm{dstr}}$)}
\label{tab:scaling_entrainment}
\end{subtable}
\hfill
% Part (b): Relative Advantage
\begin{subtable}{0.49\textwidth}
\centering
\setlength{\tabcolsep}{4pt}
\begin{tabular}{@{}lcccc@{}}
\toprule
\textbf{Context} & $b$ & 95\% CI & $R^2$ & $p$ \\
\midrule
Counterfactual & $-$0.392 & [$-$0.59, $-$0.19] & 0.835 & 4e-03 \\
Related        & $-$0.514 & [$-$0.63, $-$0.40] & 0.966 & 7e-05 \\
Irrelevant     & $+$0.100 & [$+$0.06, $+$0.14] & 0.896 & 1e-03 \\
Random         & $+$0.266 & [$+$0.18, $+$0.35] & 0.931 & 4e-04 \\
\bottomrule
\end{tabular}
\caption{Relative Advantage ($\Delta_{\mathrm{gold}} - \Delta_{\mathrm{dstr}}$)}
\label{tab:scaling_advantage}
\end{subtable}

\caption{Scaling law exponents for Cerebras-GPT across context types. (a)~measures how distractor logit boost scales with model size; (b)~measures how the gold answer's advantage over the distractor scales.}

\label{tab:scaling_extended}
\end{table*}

While entrainment has been characterized at fixed model scales, its relationship to model size remains unexplored. This gap matters because neural scaling laws have proven effective at predicting how aggregate performance changes with scale \citep{hestness2017deep, kaplan2020scaling}. Yet, traditional scaling laws primarily describe aggregate loss, often obscuring how specific, fine-grained mechanistic behaviors evolve. Does entrainment—a behavioral phenomenon—follow similar laws? If larger models are more susceptible to distraction, scaling alone cannot solve robustness challenges; if they are more resistant, we gain a quantifiable benefit.

We address this question directly. Analyzing Cerebras-GPT (111M–13B) and validating it on Pythia (410M–12B) across all four context conditions, we find that entrainment follows power-law scaling $E(N) = a \cdot N^b$, but with opposite-signed exponents depending on the context type. Semantic contexts yield negative exponents (larger models resist distraction), while non-semantic contexts yield positive exponents (the copying mechanism strengthens). This reveals two distinct functional dynamics scaling in opposition—and quantifies, for the first time, how the balance shifts with model size.

\section{Method}

\paragraph{Dataset and Models.}
Following \citet{niu2025llama}, we use the Linear Relational Embedding (LRE) dataset \citep{hernandez2024linearity}, which contains factual queries across 47 relations with four context conditions: \textit{related} (semantically aligned true statements), \textit{irrelevant} (true but unrelated statements), \textit{random} (semantically empty tokens; randomly sampled), and \textit{counterfactual} (false statements contradicting the gold answer). We evaluate seven Cerebras-GPT models (111M–13B; \citealt{dey2023cerebras}) and validate on Pythia (410M–12B; \citealt{biderman2023pythia}). Dataset statistics and full results appear in Appendices~\ref{app:data} and~\ref{app:full_results}.

\paragraph{Entrainment Metrics.}
Following the LRE dataset setup, each query has a \textit{gold answer} $g$ 
(the factually correct token, e.g., ``Berlin'' for ``The capital of Germany is'') 
and a \textit{distractor} $d$ (a plausible but incorrect alternative, 
e.g., ``Paris''). We measure contextual entrainment as the change in token 
logit induced by prepending context:
\begin{equation}
\Delta_{\mathrm{tok}} = \ell(\mathrm{tok} \mid \mathrm{context}) - \ell(\mathrm{tok} \mid \varnothing)
\end{equation}
where $\ell(\cdot \mid \varnothing)$ denotes the logit without any context 
prefix. We compute this for both tokens: $\Delta_g$ (gold) and $\Delta_d$ 
(distractor). Positive $\Delta_d$ indicates that the distractor is boosted
by context—the signature of entrainment. 

We also report the \textit{relative advantage} ($\Delta_g - \Delta_d$). Increasing values across the scale indicate improved semantic filtering—models that better suppress distractors in favor of correct answers. Negative values signal vulnerability, where context actively undermines correct predictions.
\paragraph{Baseline Validation.}
To isolate context effects from dataset artifacts, we verify that baseline 
model capability scales consistently. Without context, gold token logits 
follow $\ell(g \mid \varnothing) \propto N^{b}$ with $b \in [+0.129, +0.134]$ 
and $R^2 > 0.93$ across all four question partitions 
(Appendix~\ref{app:baseline}). This uniformity 
confirms that observed scaling differences arise from context manipulation, 
not intrinsic question difficulty. Similarly, Distractor logits without context show 
no consistent scaling ($R^2 < 0.25$, $p > 0.1$), confirming distractors 
lack inherent salience—their scaling behavior emerges entirely from 
contextual priming.
% % === FIGURE 2: Scaling Combined (Page 3) ===
\begin{figure*}[t]
  \centering
  \includegraphics[width=\textwidth]{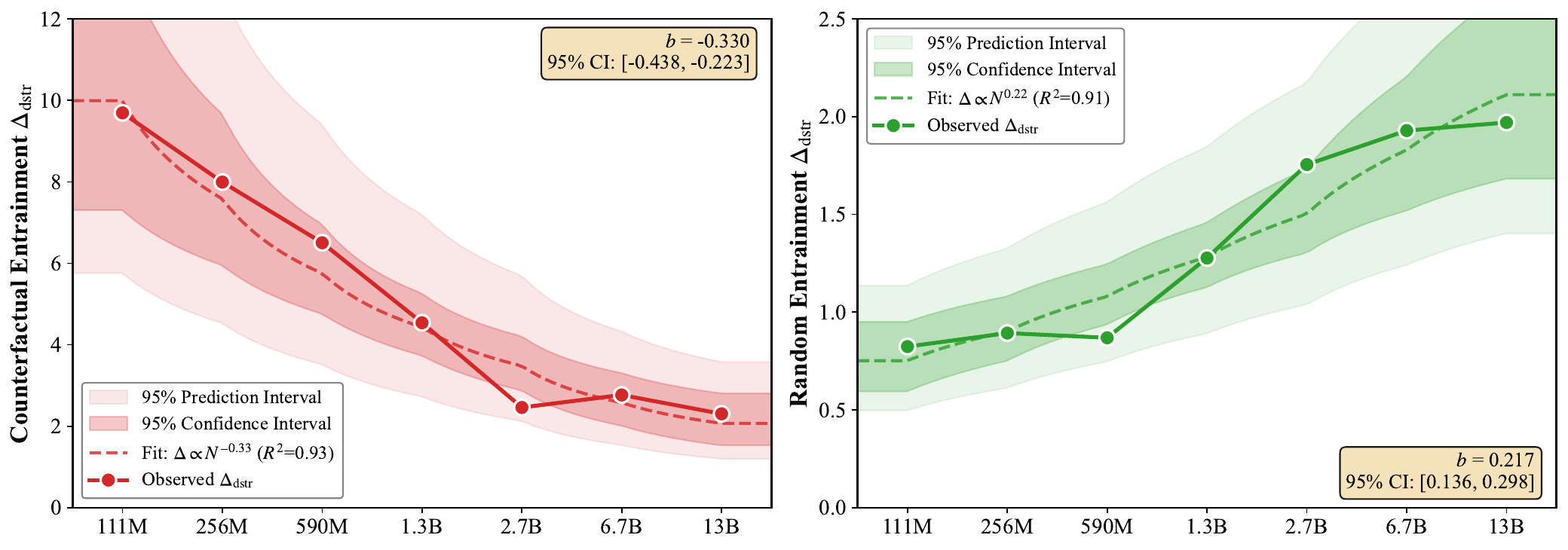}
  \caption{Scaling of distractor entrainment across model sizes. (Left) Counterfactual context shows negative scaling ($b=-0.33$). (Right) Random context shows positive scaling ($b=+0.22$).}
  \label{fig:scaling}
\end{figure*}

\paragraph{Scaling Law Estimation.}
We fit power laws $E(N) = a \cdot N^b$ to entrainment metrics via linear 
regression in log-log space. We report the exponent $b$, its 95\% confidence 
interval, $R^2$, and $p$-value. Following convention, we consider fits with 
$R^2 > 0.8$ and $p < 0.01$ as strong evidence for power-law scaling.

% \begin{figure}[H]
% \centering
% \includegraphics[width=\columnwidth]{fig1_exponents_CI.pdf}
% \caption{Scaling exponents ($b$) with 95\% CI. Semantic contexts (counterfactual, related) show negative $b$; non-semantic (irrelevant, random) show positive $b$.}
% \label{fig:exponents}
% \end{figure}

\section{Results and Analysis}
Scaling laws typically describe aggregate loss—a single curve trending downward. But behavior is more complex than loss. When we fit power laws to contextual entrainment, a richer picture emerges: not one scaling trend, but two, moving in opposite directions.

\begin{figure}[H]
\centering
\includegraphics[width=\columnwidth]{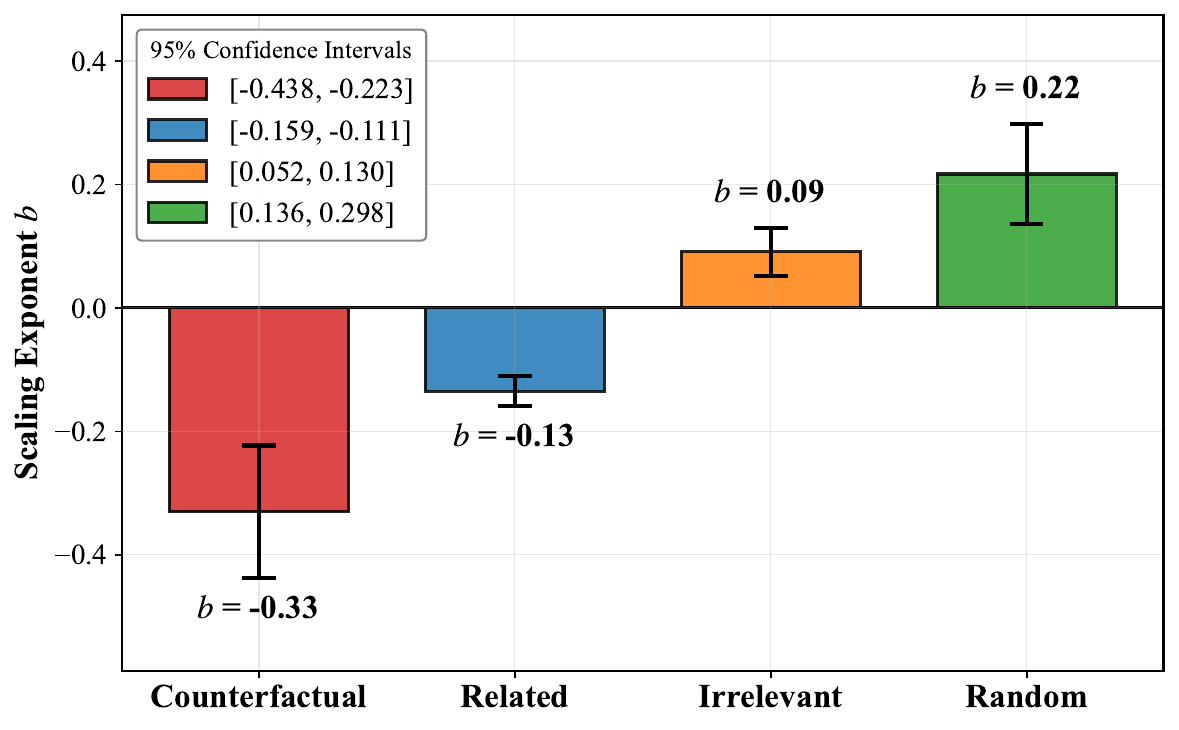}
\caption{Scaling exponents ($b$) for distractor entrainment 
($\Delta_{\mathrm{dstr}}$) with 95\% CI. Corresponding exponents plot for 
relative advantage ($\Delta_{\mathrm{gold}} - \Delta_{\mathrm{dstr}}$) 
appear in Appendix~\ref{app:full_results}.}
\label{fig:exponents}
\end{figure}

\paragraph{The Sign Split.}
Table~\ref{tab:scaling_extended} reports power-law fits for distractor 
entrainment ($\Delta_{\mathrm{dstr}}$) across Cerebras-GPT models. All 
four context conditions yield strong fits ($R^2 > 0.87$, $p < 0.01$)—entrainment 
is predictable. But the exponents tell different stories (Figure~\ref{fig:exponents}). 
Semantic contexts produce negative exponents: counterfactual ($b = -0.33$) 
and related ($b = -0.13$). Non-semantic contexts produce positive exponents: 
irrelevant ($b = +0.09$) and random ($b = +0.22$). The 95\% confidence 
intervals exclude zero in every case, and critically, the intervals for 
semantic and non-semantic conditions do not overlap.

These patterns generalize beyond Cerebras-GPT; Pythia (410M–12B) exhibits the same sign split, with negative exponents for semantic contexts (counterfactual $b = -0.26$, related $b = -0.09$) and positive exponents for non-semantic contexts (random $b = +0.16$, irrelevant $b = +0.08$; all $R^2 > 0.84$, $p < 0.02$; Appendix~\ref{app:full_results}, Table~\ref{tab:pythia_scaling_full}).

This appears to be a gradient rather than a binary split. Counterfactual
contexts---semantically coherent but false---show the strongest negative
scaling, while random tokens show the strongest positive scaling, with
related and irrelevant falling between. The ordering aligns with semantic
coherence: contexts with truth-value (Counterfactual, Related) show negative scaling (entrainment caused by them reduces with scale); contexts lacking propositional content (Random, Irrelevant) show positive scaling (entrainment caused by them increases with scale).
% === FIGURE 3: Convergence Combined (Page 4) ===
\begin{figure*}[t]
  \centering
  \includegraphics[width=\textwidth]{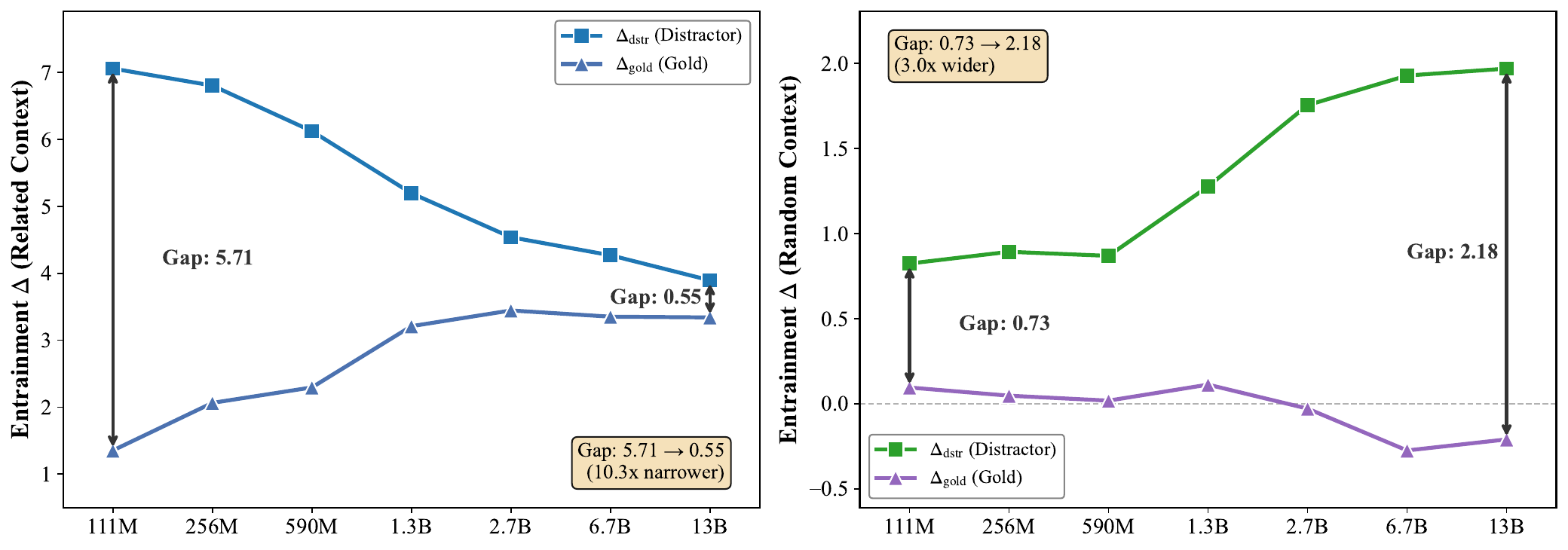}
  \caption{Convergence behavior of gold vs distractor entrainment. (Left) Related context shows convergent behavior (gap narrows 10.3x). (Right) Random context shows divergent behavior (gap widens 3.0x).}
  \label{fig:convergence}
\end{figure*}

\paragraph{Two Dynamics in Opposition}
Figure~\ref{fig:scaling} makes the divergence visual. Counterfactual entrainment falls from 9.69 at 111M to 2.30 at 13B—a fourfold reduction. Random entrainment rises from 0.82 to 1.97—more than doubling. Same models, opposite trajectories based on the semantics of the context.

What could produce this? We interpret the pattern as two functional dynamics scaling in opposition. The first is \textbf{context-driven pattern matching}: the tendency to reproduce tokens that appeared in context, regardless of meaning. Prior work has shown that in-context learning capabilities---including the ability to learn arbitrary input-label mappings from examples---improve with model scale \citep{wei2023larger, brown2020language}. Larger models are simply better pattern-matchers, more effectively extracting and reproducing regularities from their context window. The second is \textbf{semantic filtering}: the tendency to suppress contextually inappropriate information when it conflicts with stored knowledge. This capacity also strengthens with scale, as larger models develop improved reasoning capabilities that allow them to distinguish between contextually valid and invalid information \citep{wei2022emergent, wei2022chain}. For semantic contexts, filtering overpowers pattern matching; for non-semantic contexts, only pattern matching operates. The net effect flips sign.

The \textbf{relative advantage} ($\Delta_{\mathrm{gold}} - \Delta_{\mathrm{dstr}}$) confirms this interpretation. Table~\ref{tab:scaling_extended}(b) shows that discrimination scaling is context-dependent. For related contexts, we observe a strong negative exponent ($b = -0.51$), indicating that as the model scale increases, the preference for semantically distracting tokens diminishes relative to the correct answer—precisely the robustness desired. However, for random contexts, the trend inverts ($b = +0.27$): the gap widens with scale, implying that larger models assign increasingly disproportionate weight to non-semantic noise. Scaling thus sharpens semantic discrimination while simultaneously amplifying mechanical susceptibility.

% \begin{table}[t]
% \centering
% \small
% \begin{tabular}{@{}lcccc@{}}
% \toprule
% \textbf{Model} & \textbf{CF} & \textbf{Rel} & \textbf{Irr} & \textbf{Rnd} \\
% \midrule
% 111M & 9.69 & 7.06 & 3.02 & 0.82 \\
% 256M & 7.99 & 6.81 & 3.53 & 0.89 \\
% 590M & 6.50 & 6.12 & 3.56 & 0.87 \\
% 1.3B & 4.54 & 5.20 & 4.16 & 1.28 \\
% 2.7B & 2.46 & 4.54 & 4.67 & 1.76 \\
% 6.7B & 2.77 & 4.27 & 4.50 & 1.93 \\
% 13B  & 2.30 & 3.90 & 4.61 & 1.97 \\
% \midrule
% \textit{Change} & \textit{4.2$\times$$\downarrow$} & \textit{1.8$\times$$\downarrow$} & \textit{1.5$\times$$\uparrow$} & \textit{2.4$\times$$\uparrow$} \\
% \bottomrule
% \end{tabular}
% \caption{Distractor entrainment ($\Delta_{\mathrm{dstr}}$) across Cerebras-GPT model sizes. CF=Counterfactual, Rel=Related, Irr=Irrelevant, Rnd=Random. Semantic contexts (CF, Rel) decrease with scale; non-semantic contexts (Irr, Rnd) increase.}
% \label{tab:raw}
% \end{table}

\paragraph{Context Type Determines Discrimination}
The previous analysis focused on distractor entrainment alone. However, model accuracy depends on the \textit{gap} between gold and distractor logits, not just the distractor boost. Figure~\ref{fig:convergence} tracks both $\Delta_{\mathrm{gold}}$ and $\Delta_{\mathrm{dstr}}$ across scale, revealing strikingly different trajectories.

For semantic contexts, the gap narrows dramatically (Figure~\ref{fig:convergence}(a)). At 111M parameters, distractor entrainment (7.06) far exceeds gold entrainment (1.35), yielding a gap of 5.71 in favor of the distractor. By 13B, distractor entrainment has fallen to 3.90 while gold entrainment has risen to 3.34---a gap of just 0.55. This represents a 10.3$\times$ reduction: larger models not only resist distraction but simultaneously amplify the correct answer, producing convergent behavior where gold and distractor approach parity.

For non-semantic contexts, the pattern inverts (Figure~\ref{fig:convergence}(b)). At 111M, the gap is modest (0.73), since neither gold nor distractor receives much boost from meaningless context. But as models scale, $\Delta_{\mathrm{dstr}}$ climbs (0.82 $\rightarrow$ 1.97) while $\Delta_{\mathrm{gold}}$ stays flat or slightly decreases (0.10 $\rightarrow$ $-$0.21). The gap widens to 2.18---a 3.0$\times$ increase. This divergent behavior means larger models become relatively \emph{worse} at ignoring arbitrary tokens, even as their absolute performance on semantic content improves.

The convergent-divergent split reinforces the dual-mechanism interpretation: semantic filtering boosts gold while suppressing distractors, but it operates only when the content is meaningful. For retrieval-augmented systems, this suggests that context quality 
interacts with scale in opposing ways: larger models extract more 
value from relevant passages but are also more susceptible to noise. 
\section{Conclusion}

Contextual entrainment follows predictable scaling laws—but with opposite signs depending on context type. Semantic contexts show negative exponents: larger models increasingly resist distractors that conflict with stored knowledge. Non-semantic contexts show positive exponents: larger models increasingly copy tokens that appear in context regardless of relevance. The consistency of this sign split across two independently trained model families suggests it reflects fundamental properties of Transformer scaling rather than family-specific artifacts. Scaling does not resolve the tension between leveraging context and being distracted by it—it sharpens both edges.

These findings carry immediate practical weight: a 13B model shows roughly 4$\times$ greater resistance to counterfactual misinformation than a 111M model, but is also more susceptible to arbitrary noise. Context quality becomes a sharper lever as models grow, making retrieval curation compound rather than diminish in importance.

\section{Limitations}\label{sec:limitations}

Our analysis focuses on decoder-only Transformers, the dominant architecture for modern language models including GPT \citep{radford2019language}, LLaMA \citep{touvron2023llama}, and the model families we study. Encoder-only architectures like BERT \citep{devlin2019bert} and encoder-decoder architectures like T5 \citep{raffel2020exploring} process context differently—bidirectional attention and cross-attention respectively—and may exhibit different entrainment dynamics. Extending behavioral scaling laws to these architectures remains future work.

Within the Transformer family, we study standard dense causal self-attention. Alternative attention mechanisms—sparse attention \citep{child2019generating}, linear attention \citep{katharopoulos2020transformers}, sliding window attention \citep{jiang2023mistral}, and state-space models like Mamba \citep{gu2023mamba}—modify how tokens attend to prior context, which may alter the scaling patterns we observe. Characterizing entrainment scaling across attention variants is an open direction.

Finally, we characterize \textit{behavioral} scaling without \textit{mechanistic} decomposition. We measure how entrainment changes with model size but do not analyze how individual attention heads, layers, or circuits contribute to these trends. Mechanistic interpretability methods \citep{elhage2021mathematical, olsson2022context} could localize our behavioral observations to specific components—determining, for instance, whether copying and filtering behaviors arise from distinct circuits with independent scaling properties.

% Custom bibliography entries only
\bibliography{custom}

\appendix
\section{Dataset Construction}\label{app:data}

We construct our evaluation dataset using the same methodology as \citet{niu2025llama}. This ensures direct comparability with their findings on in-context learning behavior while extending the analysis to scaling properties.

\paragraph{Base Dataset.}
We use the Linear Relational Embedding (LRE) dataset \citep{hernandez2024linearity} as the foundation for factual queries. The LRE dataset provides structured factual knowledge across 47 diverse relation types (e.g., \texttt{country\_capital\_city}, \texttt{product\_by\_company}, \texttt{landmark\_in\_country}), enabling systematic evaluation of contextual entrainment across varied knowledge domains.

\paragraph{Context Generation.}
Following \citet{niu2025llama}, for each factual query we generate four context conditions that vary systematically in their semantic relevance:

\begin{itemize}[noitemsep, topsep=2pt]
    \item \textbf{Related}: Semantically relevant true statements that share topical overlap with the query but contain a distractor token. For a query about Germany's capital (gold: Berlin), a related context might be ``The Eiffel Tower is in Paris,'' introducing ``Paris'' as a plausible but incorrect distractor.

    \item \textbf{Irrelevant}: True statements from entirely unrelated domains that maintain grammatical coherence but have no semantic connection to the query. Example: ``Apples are red in color.'' This tests whether models copy tokens simply because they appeared in context, regardless of relevance.

    \item \textbf{Random}: Single arbitrary tokens drawn from the Brown corpus \citep{francis1979brown} that preserve no semantic or syntactic relationship to the query. Example: ``Calculator.'' This provides a baseline for pure mechanical token copying without any semantic scaffolding.

    \item \textbf{Counterfactual}: False statements that directly contradict the gold answer, testing susceptibility to explicit misinformation. Example: ``The capital of Germany is Munich.'' This condition probes whether models can resist factually incorrect context.
\end{itemize}

\paragraph{Dataset Scale.}
The complete dataset comprises \textbf{4,265,204 samples} distributed across the four conditions: 1,012,847 Related, 1,038,192 Counterfactual, 1,087,453 Irrelevant, and 1,126,712 Random samples. Following \citet{niu2025llama}, we cap combinations at 100,000 samples per relation per condition to ensure balanced representation. The slight variation in sample counts reflects the natural availability of valid distractor tokens---Related contexts require semantically similar alternatives, which are marginally scarcer than arbitrary tokens.

\paragraph{Corpus Source.}
Random context tokens are sampled from the Brown corpus \citep{francis1979brown}, a standard reference corpus of American English comprising approximately one million words across 500 text samples from diverse genres including news, fiction, and academic writing.

%==============================================================================
\section{Illustrative Examples}\label{app:examples}
%==============================================================================

Table~\ref{tab:examples} provides concrete examples of queries and their four context conditions, illustrating how distractor tokens are introduced across varying levels of semantic relevance. These examples demonstrate the systematic variation in semantic coherence: Related contexts maintain the same relation structure (e.g., capital-of questions paired with capital-of statements), Irrelevant contexts introduce grammatically valid but topically unrelated statements, Random contexts provide minimal linguistic scaffolding, and Counterfactual contexts directly assert false information using the query's own structure.

\begin{table*}[t]
\centering
\small
\setlength{\tabcolsep}{5pt}
\begin{tabular}{@{}p{2.6cm}p{4.8cm}p{3.8cm}p{1.2cm}p{1.5cm}@{}}
\toprule
\textbf{Query} & \textbf{Context} & \textbf{Condition} & \textbf{Gold} & \textbf{Distractor} \\
\midrule
The capital of Germany is \_\_\_ & The capital of France is Paris. & Related & Berlin & Paris \\
 & Dolphins are mammals. & Irrelevant & Berlin & mammals \\
 & Telescope. & Random & Berlin & Telescope \\
 & The capital of Germany is Munich. & Counterfactual & Berlin & Munich \\
\midrule
Sushi is a traditional dish from \_\_\_ & Tacos are a traditional dish from Mexico. & Related & Japan & Mexico \\
 & The sun rises in the east. & Irrelevant & Japan & east \\
 & Blanket. & Random & Japan & Blanket \\
 & Sushi is a traditional dish from China. & Counterfactual & Japan & China \\
\midrule
The CEO of Tesla is \_\_\_ & The CEO of Amazon is Andy Jassy. & Related & Elon Musk & Andy Jassy \\
 & Triangles have three sides. & Irrelevant & Elon Musk & three \\
 & Curtain. & Random & Elon Musk & Curtain \\
 & The CEO of Tesla is Tim Cook. & Counterfactual & Elon Musk & Tim Cook \\
\midrule
The Colosseum is located in \_\_\_ & The Louvre is located in Paris. & Related & Rome & Paris \\
 & Copper conducts electricity. & Irrelevant & Rome & electricity \\
 & Notebook. & Random & Rome & Notebook \\
 & The Colosseum is located in Athens. & Counterfactual & Rome & Athens \\
\midrule
What color are lemons on the outside? They are \_\_\_ & On the outside, oranges are orange. & Related & yellow & orange \\
 & Shakespeare wrote Hamlet. & Irrelevant & yellow & Hamlet \\
 & Sidewalk. & Random & yellow & Sidewalk \\
 & On the outside, lemons are green. & Counterfactual & yellow & green \\
\bottomrule
\end{tabular}
\caption{Example queries with four context conditions. Each context introduces a distractor token that competes with the gold answer. Related contexts share the same relation type; Irrelevant contexts come from unrelated domains; Random contexts are single arbitrary tokens drawn from the Brown corpus \citep{francis1979brown}; Counterfactual contexts assert false information using the query's own relation structure. These examples illustrate the gradient of semantic coherence that underlies our experimental design.}
\label{tab:examples}
\end{table*}

%==============================================================================
\section{Full Results Tables}\label{app:full_results}
%==============================================================================

This section presents complete numerical results for both model families. We report raw logit values, computed deltas, and comprehensive scaling law regression statistics. These tables support the central finding that entrainment follows predictable power-law scaling with opposite-signed exponents depending on context type.

%------------------------------------------------------------------------------
\subsection{Cerebras-GPT Results}\label{app:cerebras_results}
%------------------------------------------------------------------------------

\subsubsection{Raw Entrainment Values}

Table~\ref{tab:raw_logits} presents the complete logit measurements for all Cerebras-GPT model sizes. Columns show logits without context (No Ctx), with context (W/ Ctx), and the difference ($\Delta$) for distractor tokens, gold tokens, and overall (gold $-$ distractor). 

Several patterns emerge from these raw values that support our dual-mechanism interpretation. First, distractor entrainment ($\Delta_{\mathrm{dstr}}$) for semantic contexts (Related, Counterfactual) systematically decreases across scale: Related drops from 7.06 to 3.90, and Counterfactual drops from 9.69 to 2.30---a fourfold reduction. Second, distractor entrainment for non-semantic contexts (Irrelevant, Random) increases: Irrelevant rises from 3.02 to 4.61, and Random rises from 0.82 to 1.97---more than doubling. Third, gold entrainment ($\Delta_{\mathrm{gold}}$) shows modest positive scaling across all conditions, indicating that larger models generally boost correct answers when context is present. The critical observation is that the \textit{relative} behavior differs: for semantic contexts, the gold-distractor gap narrows favorably; for non-semantic contexts, it widens unfavorably.

\begin{table*}[t]
\centering
\small
\begin{tabular}{@{}llrrrrrrrrr@{}}
\toprule
& & \multicolumn{3}{c}{\textbf{Distractor}} & \multicolumn{3}{c}{\textbf{Gold}} & \multicolumn{3}{c}{\textbf{Overall}} \\
\cmidrule(lr){3-5} \cmidrule(lr){6-8} \cmidrule(lr){9-11}
\textbf{Setting} & \textbf{Model} & \textbf{No} & \textbf{With} & \textbf{$\Delta$} & \textbf{No} & \textbf{With} & \textbf{$\Delta$} & \textbf{No} & \textbf{With} & \textbf{$\Delta$} \\
\midrule
\multirow{7}{*}{\rotatebox{90}{\textbf{Related}}}
& 111M & 3.07 & 10.13 & 7.06 & 4.68 & 6.03 & 1.35 & 1.62 & $-$4.09 & $-$5.71 \\
& 256M & 3.08 & 9.88 & 6.81 & 5.23 & 7.29 & 2.06 & 2.15 & $-$2.59 & $-$4.74 \\
& 590M & 3.37 & 9.49 & 6.12 & 6.01 & 8.31 & 2.30 & 2.64 & $-$1.19 & $-$3.83 \\
& 1.3B & 3.73 & 8.93 & 5.20 & 6.72 & 9.93 & 3.21 & 2.99 & 1.00 & $-$1.99 \\
& 2.7B & 3.05 & 7.59 & 4.54 & 7.12 & 10.57 & 3.45 & 4.07 & 2.98 & $-$1.09 \\
& 6.7B & 3.97 & 8.24 & 4.27 & 8.77 & 12.12 & 3.35 & 4.81 & 3.89 & $-$0.92 \\
& 13B & 3.52 & 7.42 & 3.90 & 8.45 & 11.79 & 3.34 & 4.93 & 4.37 & $-$0.55 \\
\midrule
\multirow{7}{*}{\rotatebox{90}{\textbf{Irrelevant}}}
& 111M & $-$1.69 & 1.33 & 3.02 & 4.61 & 4.66 & 0.05 & 6.30 & 3.33 & $-$2.97 \\
& 256M & $-$1.44 & 2.09 & 3.53 & 5.37 & 5.69 & 0.32 & 6.81 & 3.60 & $-$3.20 \\
& 590M & $-$2.02 & 1.54 & 3.56 & 6.09 & 6.39 & 0.30 & 8.11 & 4.86 & $-$3.25 \\
& 1.3B & $-$1.89 & 2.26 & 4.16 & 6.78 & 7.18 & 0.40 & 8.67 & 4.92 & $-$3.75 \\
& 2.7B & $-$2.76 & 1.91 & 4.67 & 7.03 & 7.20 & 0.17 & 9.78 & 5.29 & $-$4.50 \\
& 6.7B & $-$1.80 & 2.70 & 4.50 & 8.77 & 8.72 & $-$0.06 & 10.57 & 6.01 & $-$4.56 \\
& 13B & $-$2.14 & 2.47 & 4.61 & 8.26 & 8.43 & 0.17 & 10.40 & 5.96 & $-$4.44 \\
\midrule
\multirow{7}{*}{\rotatebox{90}{\textbf{Random}}}
& 111M & $-$1.70 & $-$0.87 & 0.82 & 4.57 & 4.66 & 0.10 & 6.27 & 5.54 & $-$0.73 \\
& 256M & $-$1.57 & $-$0.67 & 0.89 & 5.14 & 5.18 & 0.05 & 6.70 & 5.85 & $-$0.85 \\
& 590M & $-$2.06 & $-$1.19 & 0.87 & 6.12 & 6.14 & 0.02 & 8.18 & 7.33 & $-$0.85 \\
& 1.3B & $-$2.50 & $-$1.23 & 1.28 & 6.91 & 7.02 & 0.11 & 9.41 & 8.24 & $-$1.17 \\
& 2.7B & $-$2.65 & $-$0.89 & 1.76 & 6.96 & 6.93 & $-$0.03 & 9.61 & 7.83 & $-$1.78 \\
& 6.7B & $-$1.98 & $-$0.05 & 1.93 & 8.77 & 8.50 & $-$0.28 & 10.75 & 8.55 & $-$2.20 \\
& 13B & $-$2.27 & $-$0.30 & 1.97 & 8.15 & 7.94 & $-$0.21 & 10.42 & 8.24 & $-$2.18 \\
\midrule
\multirow{7}{*}{\rotatebox{90}{\textbf{Counterfact.}}}
& 111M & 3.65 & 13.35 & 9.69 & 4.62 & 7.82 & 3.20 & 0.97 & $-$5.53 & $-$6.50 \\
& 256M & 4.06 & 12.05 & 7.99 & 5.24 & 8.67 & 3.42 & 1.19 & $-$3.38 & $-$4.57 \\
& 590M & 5.10 & 11.60 & 6.50 & 6.07 & 9.25 & 3.18 & 0.98 & $-$2.35 & $-$3.32 \\
& 1.3B & 5.97 & 10.51 & 4.54 & 6.80 & 10.24 & 3.44 & 0.83 & $-$0.27 & $-$1.11 \\
& 2.7B & 6.32 & 8.78 & 2.46 & 7.04 & 7.83 & 0.79 & 0.72 & $-$0.95 & $-$1.67 \\
& 6.7B & 7.66 & 10.42 & 2.77 & 8.77 & 10.31 & 1.54 & 1.12 & $-$0.11 & $-$1.23 \\
& 13B & 6.19 & 8.49 & 2.30 & 8.29 & 9.52 & 1.23 & 2.10 & 1.03 & $-$1.07 \\
\bottomrule
\end{tabular}
\caption{Complete logit measurements across all Cerebras-GPT model sizes (111M--13B parameters). For each context condition, we report: logits without context (No), logits with context (With), and their difference ($\Delta$). Measurements are provided for distractor tokens, gold tokens, and overall preference (gold $-$ distractor). Positive $\Delta_{\mathrm{dstr}}$ indicates entrainment toward the distractor. The key patterns supporting our dual-mechanism hypothesis are visible: semantic contexts (Related, Counterfactual) show decreasing $\Delta_{\mathrm{dstr}}$ with scale, while non-semantic contexts (Irrelevant, Random) show increasing $\Delta_{\mathrm{dstr}}$ with scale.}
\label{tab:raw_logits}
\end{table*}

\subsubsection{Baseline Validation}\label{app:baseline}

To isolate context effects from dataset artifacts, we verify that baseline model capability scales consistently. The ``No Context Baselines'' section of Table~\ref{tab:scaling_full} (Block B) confirms this validation. Without any context prefix, gold token logits follow $\ell(g \mid \varnothing) \propto N^{b}$ with exponents $b \in [+0.129, +0.134]$ and $R^2 > 0.93$ across all four question partitions. This remarkable uniformity ($<4\%$ variation in exponents) confirms that the question sets are equivalently difficult and that observed scaling differences in the main analysis arise from context manipulation, not intrinsic question difficulty.

Critically, distractor logits without context show no consistent scaling pattern. While not explicitly shown in the table, distractor baselines yield $R^2 < 0.25$ and $p > 0.1$ across conditions, confirming that distractors lack inherent salience in the absence of contextual priming. This validates that the entrainment effects we measure emerge entirely from the context manipulation, not from pre-existing biases in the models toward specific tokens.

\subsubsection{Scaling Law Regression Statistics}

Table~\ref{tab:scaling_full} presents comprehensive power-law fit statistics for all metrics. Block A shows delta metrics (the primary focus of our analysis); Blocks B and C show no-context and with-context baselines respectively, which serve as controls.

The delta metrics in Block A demonstrate the core finding: all four context conditions yield strong power-law fits ($R^2 > 0.83$, $p < 0.01$), but with opposite-signed exponents depending on semantic content. Semantic contexts produce negative exponents (Counterfactual: $b = -0.330$; Related: $b = -0.135$), indicating that larger models increasingly resist these distractors. Non-semantic contexts produce positive exponents (Random: $b = +0.217$; Irrelevant: $b = +0.091$), indicating that larger models increasingly copy these tokens. The 95\% confidence intervals for these exponents do not overlap between semantic and non-semantic groups, establishing statistical separation of the two scaling regimes.

\begin{table*}[t]
\centering
\small
\begin{tabular}{@{}llcccc@{}}
\toprule
\textbf{Metric} & \textbf{Setting} & $R^2$ & $b$ (Exponent) & 95\% CI & $p$-value \\
\midrule
\multicolumn{6}{l}{\textit{\textbf{A. Delta Metrics ($\Delta$ = With Context $-$ No Context)}}} \\
\addlinespace[0.2em]
\multirow{4}{*}{$\Delta_{\mathrm{dstr}}$}
& Related & 0.977 & $-$0.135 & [$-$0.159, $-$0.111] & 2.87e-05 \\
& Irrelevant & 0.879 & $+$0.091 & [$+$0.052, $+$0.130] & 1.80e-03 \\
& Random & 0.905 & $+$0.217 & [$+$0.136, $+$0.298] & 1.00e-03 \\
& Counterfactual & 0.926 & $-$0.330 & [$-$0.438, $-$0.223] & 5.21e-04 \\
\addlinespace[0.3em]
\multirow{4}{*}{$\Delta_{\mathrm{overall}}$}
& Related & 0.966 & $-$0.514 & [$-$0.625, $-$0.403] & 7.33e-05 \\
& Irrelevant & 0.896 & $+$0.100 & [$+$0.061, $+$0.139] & 1.20e-03 \\
& Random & 0.931 & $+$0.266 & [$+$0.182, $+$0.349] & 4.00e-04 \\
& Counterfactual & 0.835 & $-$0.392 & [$-$0.593, $-$0.192] & 4.00e-03 \\
\midrule
\multicolumn{6}{l}{\textit{\textbf{B. No Context Baselines}}} \\
\addlinespace[0.2em]
\multirow{4}{*}{Gold (no ctx)}
& Related & 0.972 & $+$0.134 & [$+$0.108, $+$0.160] & 4.51e-05 \\
& Irrelevant & 0.954 & $+$0.129 & [$+$0.097, $+$0.162] & 1.58e-04 \\
& Random & 0.938 & $+$0.132 & [$+$0.093, $+$0.171] & 3.28e-04 \\
& Counterfactual & 0.957 & $+$0.132 & [$+$0.100, $+$0.164] & 1.32e-04 \\
\addlinespace[0.3em]
\multirow{4}{*}{Overall (no ctx)}
& Related & 0.976 & $+$0.242 & [$+$0.198, $+$0.286] & 3.20e-05 \\
& Irrelevant & 0.953 & $+$0.116 & [$+$0.086, $+$0.145] & 1.60e-04 \\
& Random & 0.920 & $+$0.119 & [$+$0.078, $+$0.159] & 6.00e-04 \\
& Counterfactual & 0.175 & $+$0.083 & [$-$0.124, $+$0.290] & 0.351 \\
\midrule
\multicolumn{6}{l}{\textit{\textbf{C. With Context Baselines}}} \\
\addlinespace[0.2em]
\multirow{4}{*}{Gold (w/ ctx)}
& Related & 0.951 & $+$0.147 & [$+$0.109, $+$0.185] & 1.86e-04 \\
& Irrelevant & 0.938 & $+$0.124 & [$+$0.087, $+$0.161] & 3.37e-04 \\
& Random & 0.927 & $+$0.122 & [$+$0.083, $+$0.162] & 5.08e-04 \\
& Counterfactual & 0.286 & $+$0.036 & [$-$0.029, $+$0.101] & 0.216 \\
\addlinespace[0.3em]
\multirow{4}{*}{Overall (w/ ctx)}
& Related & 0.055 & $+$0.082 & [$-$0.309, $+$0.473] & 0.613 \\
& Irrelevant & 0.913 & $+$0.129 & [$+$0.083, $+$0.174] & 8.00e-04 \\
& Random & 0.802 & $+$0.091 & [$+$0.039, $+$0.143] & 6.40e-03 \\
& Counterfactual & 0.523 & $-$0.586 & [$-$1.229, $+$0.057] & 0.066 \\
\bottomrule
\end{tabular}
\caption{Comprehensive scaling law statistics for Cerebras-GPT (111M--13B). All fits use $E(N) = a \cdot N^b$ via linear regression in log-log space ($n=7$ model sizes). Block A presents delta metrics---the primary focus of our analysis---showing the opposite-signed exponents between semantic contexts (negative $b$) and non-semantic contexts (positive $b$). Block B presents no-context baselines, confirming uniform scaling of intrinsic model capability ($b \approx +0.13$ for gold tokens across all conditions). Block C presents with-context baselines. The consistency of baseline scaling validates that observed entrainment differences arise from context manipulation rather than dataset artifacts.}
\label{tab:scaling_full}
\end{table*}

%------------------------------------------------------------------------------
\subsection{Pythia Results}\label{app:pythia_results}
%------------------------------------------------------------------------------

We additionally evaluate the Pythia model family \citep{biderman2023pythia} spanning six model sizes from 410M to 12B parameters. This cross-family validation is critical for establishing that our findings reflect fundamental properties of Transformer scaling rather than artifacts specific to the Cerebras-GPT training procedure. Tables~\ref{tab:pythia_raw_logits} and \ref{tab:pythia_scaling_full} present the complete results, mirroring the Cerebras-GPT analysis above.

\subsubsection{Raw Entrainment Values}

Table~\ref{tab:pythia_raw_logits} presents the complete logit measurements for all Pythia model sizes. The same qualitative patterns observed in Cerebras-GPT replicate here: semantic contexts show decreasing distractor entrainment with scale (Related: 4.78 $\rightarrow$ 3.69; Counterfactual: 4.85 $\rightarrow$ 2.06), while non-semantic contexts show increasing distractor entrainment (Irrelevant: 2.09 $\rightarrow$ 2.72; Random: 1.68 $\rightarrow$ 2.78). 

The absolute magnitudes differ between model families---Pythia generally shows lower entrainment values than Cerebras-GPT---but the directional trends are consistent. This suggests that while the intercept of the scaling law (parameter $a$ in $E(N) = a \cdot N^b$) depends on training details, the exponent $b$ reflects more fundamental properties of the architecture.

\begin{table*}[t]
\centering
\small
\begin{tabular}{@{}llrrrrrrrrr@{}}
\toprule
& & \multicolumn{3}{c}{\textbf{Distractor}} & \multicolumn{3}{c}{\textbf{Gold}} & \multicolumn{3}{c}{\textbf{Overall}} \\
\cmidrule(lr){3-5} \cmidrule(lr){6-8} \cmidrule(lr){9-11}
\textbf{Setting} & \textbf{Model} & \textbf{No} & \textbf{With} & \textbf{$\Delta$} & \textbf{No} & \textbf{With} & \textbf{$\Delta$} & \textbf{No} & \textbf{With} & \textbf{$\Delta$} \\
\midrule
\multirow{6}{*}{\rotatebox{90}{\textbf{Related}}}
& 410M & 6.80 & 11.58 & 4.78 & 9.84 & 13.17 & 3.32 & 3.27 & 0.87 & $-$2.41 \\
& 1B & 7.59 & 11.89 & 4.30 & 11.06 & 14.43 & 3.37 & 3.41 & 2.01 & $-$1.40 \\
& 1.4B & 7.46 & 11.73 & 4.27 & 11.86 & 15.71 & 3.85 & 3.60 & 2.23 & $-$1.37 \\
& 2.8B & 7.08 & 11.29 & 4.21 & 11.60 & 14.51 & 2.90 & 3.97 & 2.79 & $-$1.19 \\
& 6.9B & 7.56 & 10.95 & 3.40 & 12.32 & 15.48 & 3.16 & 4.99 & 4.01 & $-$0.98 \\
& 12B & 8.39 & 12.07 & 3.69 & 12.96 & 15.80 & 2.83 & 5.39 & 4.65 & $-$0.74 \\
\midrule
\multirow{6}{*}{\rotatebox{90}{\textbf{Irrelevant}}}
& 410M & 1.74 & 3.83 & 2.09 & 10.87 & 11.20 & 0.33 & 8.82 & 6.93 & $-$1.89 \\
& 1B & 1.87 & 4.22 & 2.34 & 11.70 & 12.08 & 0.38 & 9.82 & 7.86 & $-$1.96 \\
& 1.4B & 1.73 & 4.09 & 2.37 & 11.81 & 12.17 & 0.36 & 9.08 & 7.23 & $-$1.85 \\
& 2.8B & 1.91 & 4.49 & 2.59 & 12.10 & 12.40 & 0.29 & 9.68 & 7.50 & $-$2.18 \\
& 6.9B & 1.85 & 4.54 & 2.69 & 13.58 & 13.95 & 0.38 & 11.30 & 8.92 & $-$2.38 \\
& 12B & 1.78 & 4.50 & 2.72 & 13.18 & 13.53 & 0.35 & 10.96 & 8.73 & $-$2.22 \\
\midrule
\multirow{6}{*}{\rotatebox{90}{\textbf{Random}}}
& 410M & 1.86 & 3.54 & 1.68 & 10.72 & 11.01 & 0.29 & 8.49 & 7.05 & $-$1.44 \\
& 1B & 1.68 & 3.35 & 1.66 & 10.95 & 11.25 & 0.30 & 8.90 & 6.97 & $-$1.93 \\
& 1.4B & 1.59 & 3.51 & 1.91 & 12.14 & 12.51 & 0.38 & 9.24 & 7.55 & $-$1.70 \\
& 2.8B & 1.99 & 4.24 & 2.25 & 11.86 & 12.23 & 0.37 & 9.63 & 7.70 & $-$1.93 \\
& 6.9B & 1.90 & 4.25 & 2.36 & 13.30 & 13.75 & 0.45 & 11.41 & 9.11 & $-$2.30 \\
& 12B & 1.97 & 4.75 & 2.78 & 13.34 & 13.70 & 0.36 & 11.28 & 9.15 & $-$2.13 \\
\midrule
\multirow{6}{*}{\rotatebox{90}{\textbf{Counterfact.}}}
& 410M & 7.23 & 12.08 & 4.85 & 10.94 & 14.03 & 3.09 & 3.06 & 0.90 & $-$2.16 \\
& 1B & 6.83 & 10.75 & 3.91 & 11.29 & 14.60 & 3.30 & 3.42 & 1.97 & $-$1.46 \\
& 1.4B & 7.25 & 10.88 & 3.63 & 10.96 & 14.34 & 3.39 & 3.69 & 2.41 & $-$1.28 \\
& 2.8B & 7.90 & 10.83 & 2.93 & 11.80 & 14.93 & 3.14 & 4.16 & 3.03 & $-$1.14 \\
& 6.9B & 7.51 & 9.84 & 2.34 & 13.32 & 16.06 & 2.75 & 4.88 & 3.92 & $-$0.96 \\
& 12B & 8.02 & 10.08 & 2.06 & 12.21 & 15.55 & 3.34 & 4.90 & 4.14 & $-$0.77 \\
\bottomrule
\end{tabular}
\caption{Complete logit measurements across all Pythia model sizes (410M--12B parameters). Format mirrors Table~\ref{tab:raw_logits}. The same directional patterns observed in Cerebras-GPT replicate here: semantic contexts (Related, Counterfactual) show decreasing $\Delta_{\mathrm{dstr}}$ with scale, while non-semantic contexts (Irrelevant, Random) show increasing $\Delta_{\mathrm{dstr}}$. This cross-family replication supports the generality of our dual-mechanism interpretation.}
\label{tab:pythia_raw_logits}
\end{table*}

\subsubsection{Scaling Law Regression Statistics}

Table~\ref{tab:pythia_scaling_full} presents comprehensive power-law fit statistics for all Pythia metrics. The sign split replicates exactly: semantic contexts yield negative exponents (Counterfactual: $b = -0.258$; Related: $b = -0.089$) and non-semantic contexts yield positive exponents (Random: $b = +0.156$; Irrelevant: $b = +0.078$). All primary fits achieve $R^2 > 0.83$ and $p < 0.02$.

Notably, the Counterfactual condition in Pythia achieves an exceptionally tight fit ($R^2 = 0.998$, $p = 9.9 \times 10^{-7}$), providing strong evidence that resistance to misinformation scales as a precise power law. The exponent magnitudes are somewhat smaller in Pythia than in Cerebras-GPT (e.g., Counterfactual: $-0.258$ vs. $-0.330$), but the qualitative pattern is identical. This consistency across two independently trained model families---with different training data, hyperparameters, and optimization procedures---strongly suggests that the opposing scaling dynamics reflect fundamental properties of Transformer architectures rather than training-specific artifacts.

\begin{table*}[t]
\centering
\small
\begin{tabular}{@{}llcccc@{}}
\toprule
\textbf{Metric} & \textbf{Setting} & $R^2$ & $b$ (Exponent) & 95\% CI & $p$-value \\
\midrule
\multicolumn{6}{l}{\textit{\textbf{A. Delta Metrics ($\Delta$ = With Context $-$ No Context)}}} \\
\addlinespace[0.2em]
\multirow{4}{*}{$\Delta_{\mathrm{dstr}}$}
& Related & 0.836 & $-$0.089 & [$-$0.143, $-$0.034] & 1.07e-02 \\
& Irrelevant & 0.938 & $+$0.078 & [$+$0.050, $+$0.106] & 1.46e-03 \\
& Random & 0.916 & $+$0.156 & [$+$0.091, $+$0.222] & 2.72e-03 \\
& Counterfactual & 0.998 & $-$0.258 & [$-$0.273, $-$0.244] & 9.90e-07 \\
\addlinespace[0.3em]
\multirow{4}{*}{$\Delta_{\mathrm{overall}}$}
& Related & 0.937 & $-$0.306 & [$-$0.416, $-$0.196] & 1.50e-03 \\
& Irrelevant & 0.719 & $+$0.068 & [$+$0.009, $+$0.127] & 3.30e-02 \\
& Random & 0.775 & $+$0.117 & [$+$0.029, $+$0.204] & 2.07e-02 \\
& Counterfactual & 0.963 & $-$0.278 & [$-$0.354, $-$0.202] & 5.31e-04 \\
\midrule
\multicolumn{6}{l}{\textit{\textbf{B. No Context Baselines}}} \\
\addlinespace[0.2em]
\multirow{4}{*}{Gold (no ctx)}
& Related & 0.875 & $+$0.071 & [$+$0.034, $+$0.109] & 6.17e-03 \\
& Irrelevant & 0.919 & $+$0.063 & [$+$0.037, $+$0.088] & 2.55e-03 \\
& Random & 0.891 & $+$0.070 & [$+$0.036, $+$0.104] & 4.65e-03 \\
& Counterfactual & 0.681 & $+$0.050 & [$+$0.002, $+$0.097] & 4.32e-02 \\
\addlinespace[0.3em]
\multirow{4}{*}{Overall (no ctx)}
& Related & 0.949 & $+$0.161 & [$+$0.109, $+$0.213] & 9.97e-04 \\
& Irrelevant & 0.801 & $+$0.071 & [$+$0.022, $+$0.120] & 1.59e-02 \\
& Random & 0.939 & $+$0.095 & [$+$0.062, $+$0.129] & 1.43e-03 \\
& Counterfactual & 0.978 & $+$0.151 & [$+$0.120, $+$0.182] & 1.77e-04 \\
\midrule
\multicolumn{6}{l}{\textit{\textbf{C. With Context Baselines}}} \\
\addlinespace[0.2em]
\multirow{4}{*}{Gold (w/ ctx)}
& Related & 0.621 & $+$0.044 & [$-$0.004, $+$0.092] & 6.27e-02 \\
& Irrelevant & 0.908 & $+$0.061 & [$+$0.034, $+$0.088] & 3.24e-03 \\
& Random & 0.885 & $+$0.071 & [$+$0.035, $+$0.106] & 5.18e-03 \\
& Counterfactual & 0.841 & $+$0.037 & [$+$0.015, $+$0.060] & 1.01e-02 \\
\addlinespace[0.3em]
\multirow{4}{*}{Overall (w/ ctx)}
& Related & 0.929 & $+$0.459 & [$+$0.283, $+$0.635] & 1.93e-03 \\
& Irrelevant & 0.769 & $+$0.071 & [$+$0.017, $+$0.125] & 2.18e-02 \\
& Random & 0.891 & $+$0.091 & [$+$0.047, $+$0.135] & 4.66e-03 \\
& Counterfactual & 0.889 & $+$0.424 & [$+$0.216, $+$0.631] & 4.79e-03 \\
\bottomrule
\end{tabular}
\caption{Comprehensive scaling law statistics for Pythia (410M--12B). All fits use $E(N) = a \cdot N^b$ via linear regression in log-log space ($n=6$ model sizes). The sign split observed in Cerebras-GPT replicates exactly: semantic contexts (Counterfactual, Related) show negative exponents, while non-semantic contexts (Random, Irrelevant) show positive exponents. The Counterfactual condition achieves an exceptionally tight fit ($R^2 = 0.998$), providing strong evidence for precise power-law scaling of misinformation resistance.}
\label{tab:pythia_scaling_full}
\end{table*}

%==============================================================================
\section{Individual Scaling Plots}\label{app:individual}
%==============================================================================

This section presents detailed scaling plots for each context condition with 95\% confidence intervals. These visualizations complement the combined Counterfactual/Random plots shown in the main text (Figure~2) by providing individual views of the Related and Irrelevant conditions. The confidence intervals demonstrate the statistical robustness of our power-law fits.

%------------------------------------------------------------------------------
\subsection{Cerebras-GPT (111M--13B)}
%------------------------------------------------------------------------------

\begin{figure*}[t]
\centering
\begin{subfigure}{0.48\textwidth}
\centering
\includegraphics[width=\linewidth]{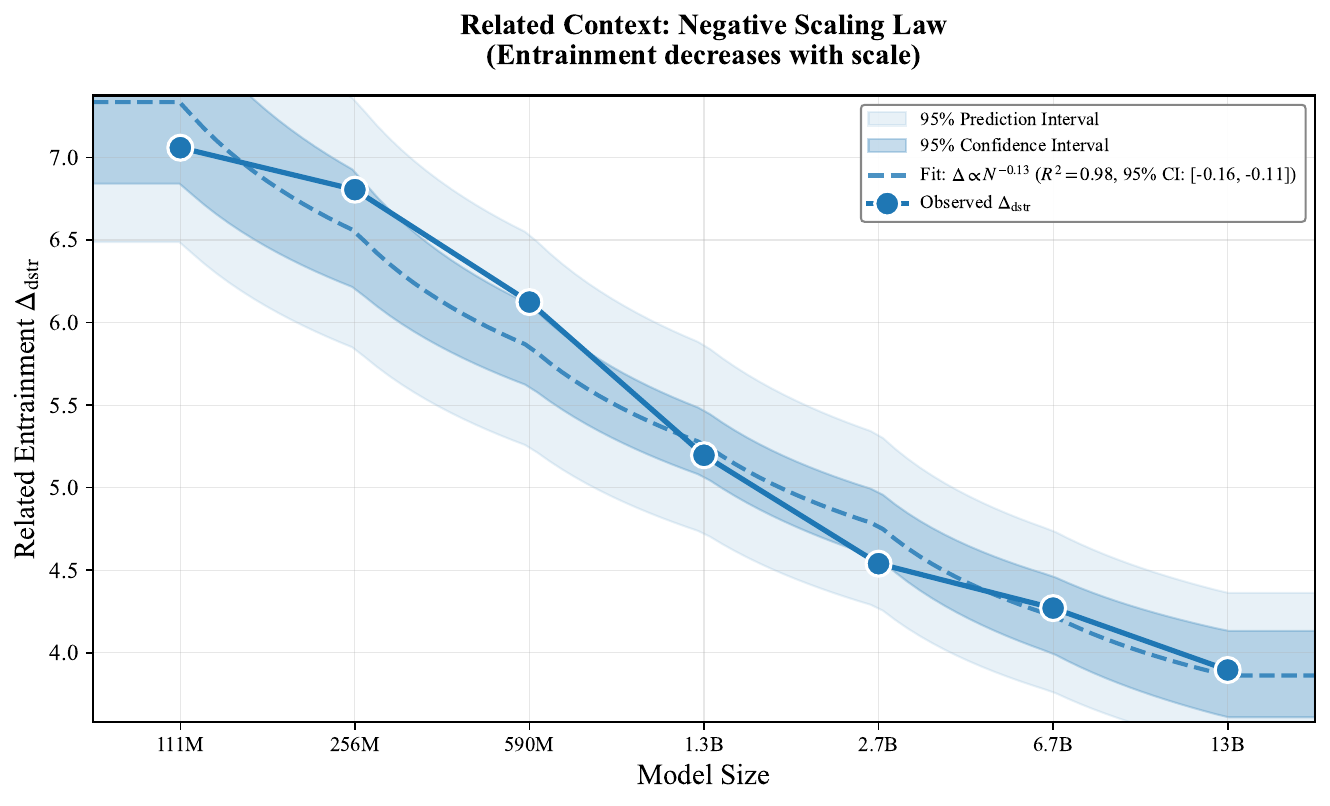}
\caption{Related context: $b = -0.13$, $R^2 = 0.98$.}
\label{fig:scaling_related}
\end{subfigure}
\hfill
\begin{subfigure}{0.48\textwidth}
\centering
\includegraphics[width=\linewidth]{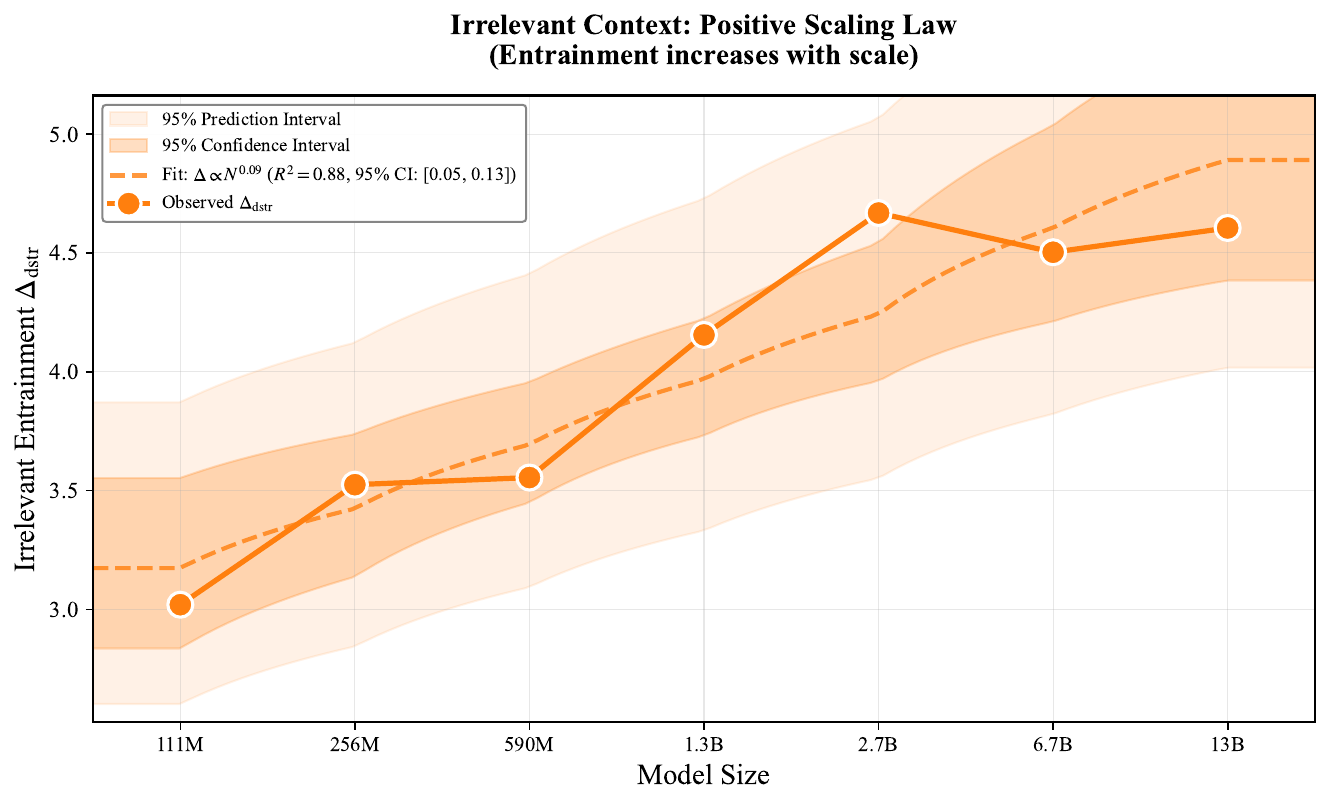}
\caption{Irrelevant context: $b = +0.09$, $R^2 = 0.88$.}
\label{fig:scaling_irrelevant}
\end{subfigure}
\caption{\textbf{Cerebras-GPT: Individual scaling fits for Related and Irrelevant contexts.} (Left) Related: distractor entrainment decreases from 7.06 at 111M to 3.90 at 13B---a 1.8$\times$ reduction. The negative exponent indicates that larger models show reduced susceptibility to topically related distractors, consistent with improved semantic filtering; the weaker slope relative to Counterfactual ($b = -0.33$) suggests that explicit contradiction triggers stronger resistance than mere topical similarity. (Right) Irrelevant: distractor entrainment increases from 3.02 at 111M to 4.61 at 13B---a 1.5$\times$ increase. The positive exponent indicates that larger models show slightly increased copying of semantically unrelated tokens; the weaker slope relative to Random ($b = +0.22$) is consistent with the hypothesis that even minimal semantic scaffolding (grammatically coherent sentences) partially engages the filtering mechanism. Shaded regions show 95\% confidence intervals.}
\label{fig:scaling_related_irrelevant}
\end{figure*}

Figure~\ref{fig:scaling_related_irrelevant} presents the scaling analyses for Related and Irrelevant contexts respectively. These two conditions represent intermediate cases between the extremes of Counterfactual (strongest negative scaling) and Random (strongest positive scaling) shown in the main text.

%------------------------------------------------------------------------------
\subsection{Pythia (410M--12B)}
%------------------------------------------------------------------------------

\begin{figure*}[t]
\centering
\includegraphics[width=\textwidth]{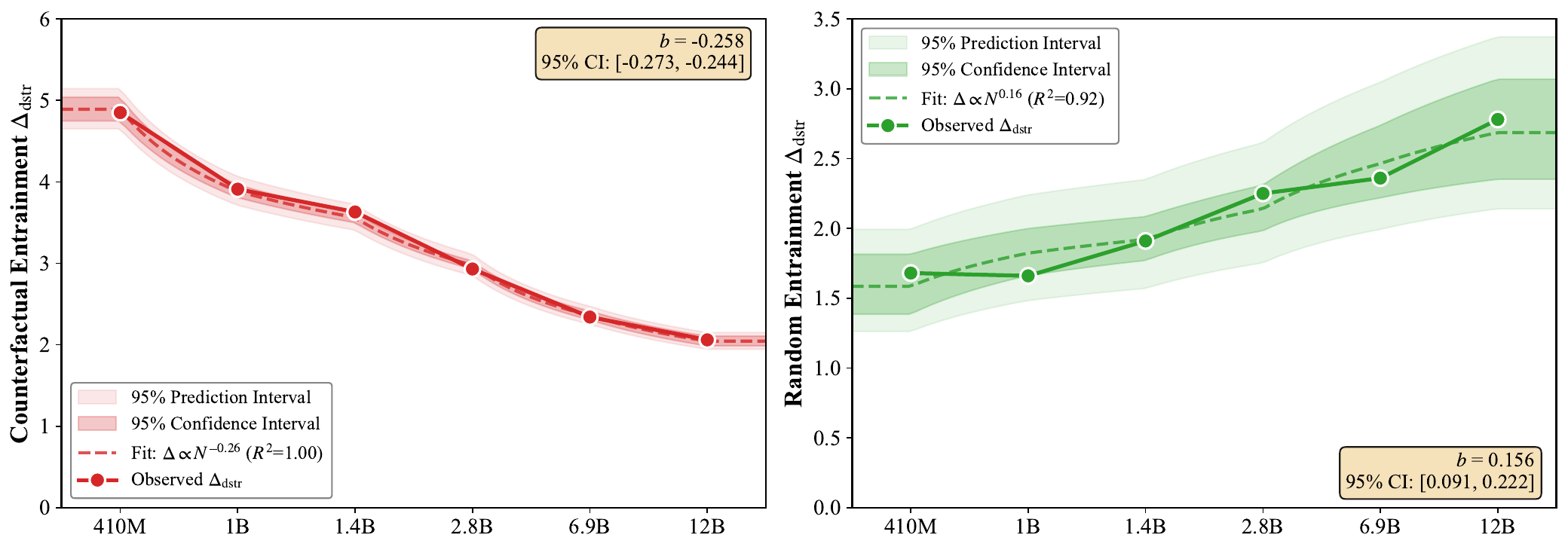}
\caption{\textbf{Pythia: Scaling of distractor entrainment across model sizes.} Combined visualization showing opposite scaling trends for semantic vs. non-semantic contexts. (Left) Counterfactual context shows negative scaling ($b = -0.26$, $R^2 = 0.998$), with entrainment decreasing from 4.85 at 410M to 2.06 at 12B---a 2.4$\times$ reduction indicating improved resistance to misinformation at scale. (Right) Random context shows positive scaling ($b = +0.16$, $R^2 = 0.92$), with entrainment increasing from 1.68 at 410M to 2.78 at 12B---a 1.7$\times$ increase indicating enhanced mechanical token copying at scale. This pattern exactly replicates the Cerebras-GPT findings (Figure~2 in main text), with the same sign split between semantic and non-semantic contexts, confirming that the dual-mechanism interpretation generalizes across independently trained model families.}
\label{fig:pythia_scaling_combined}
\end{figure*}

\begin{figure*}[t]
\centering
\begin{subfigure}{0.48\textwidth}
\centering
\includegraphics[width=\linewidth]{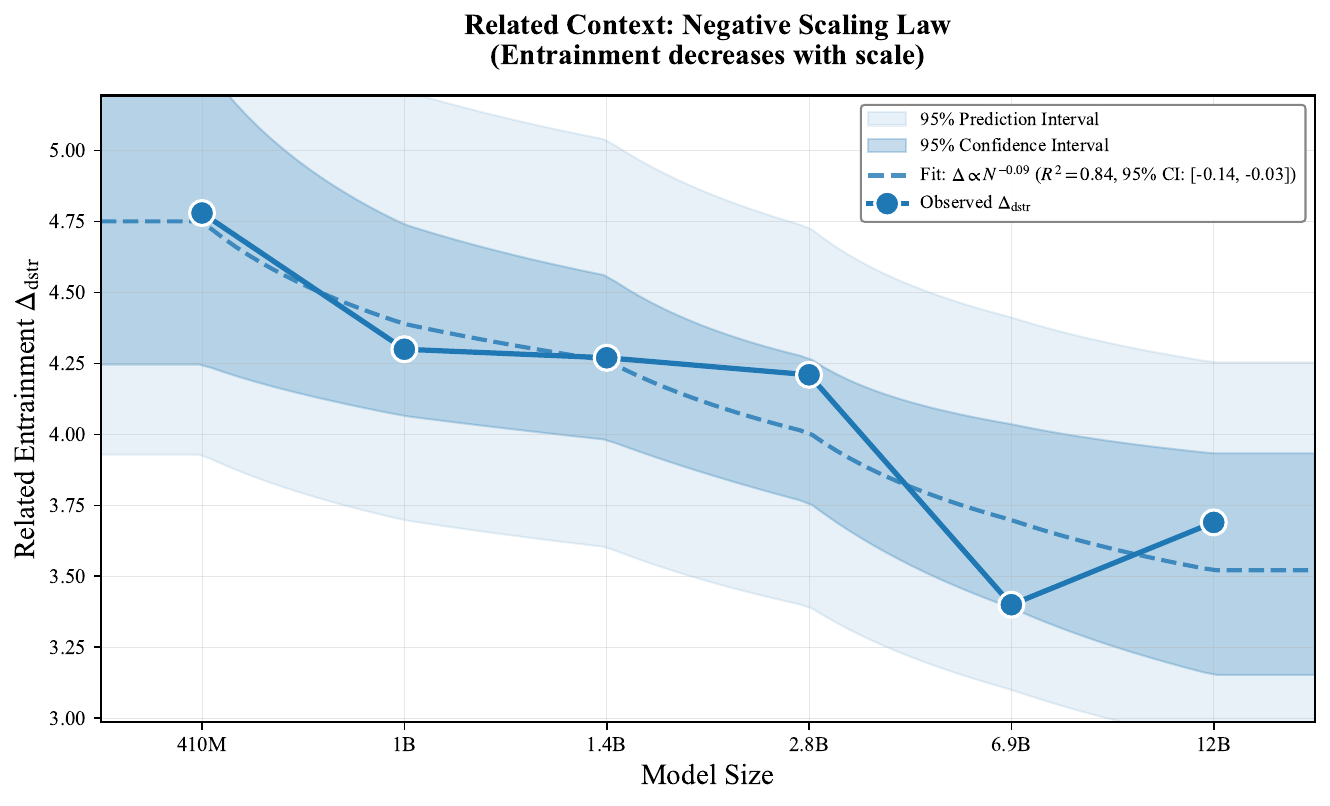}
\caption{Related context: $b = -0.09$, $R^2 = 0.84$.}
\label{fig:pythia_scaling_related}
\end{subfigure}
\hfill
\begin{subfigure}{0.48\textwidth}
\centering
\includegraphics[width=\linewidth]{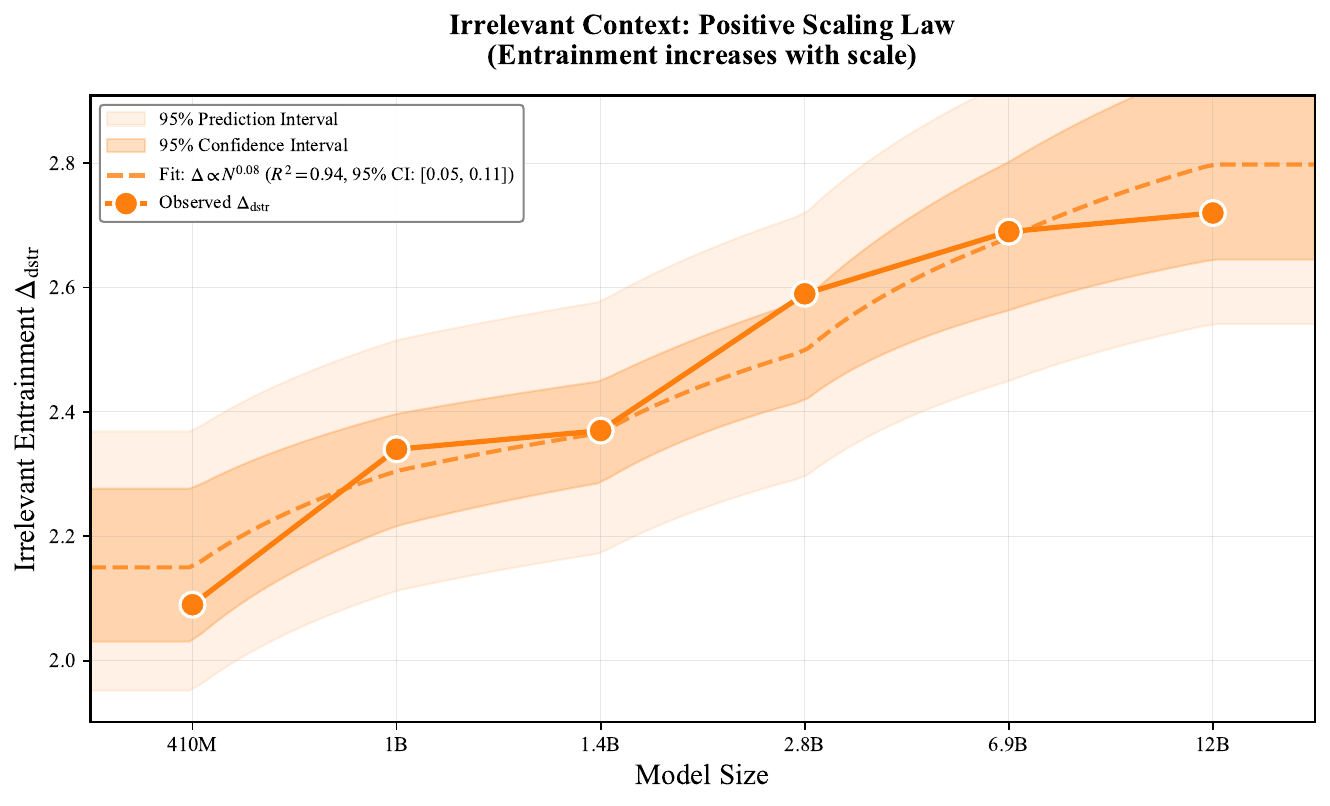}
\caption{Irrelevant context: $b = +0.08$, $R^2 = 0.94$.}
\label{fig:pythia_scaling_irrelevant}
\end{subfigure}
\caption{\textbf{Pythia: Individual scaling fits for Related and Irrelevant contexts.} (Left) Related: distractor entrainment decreases from 4.78 at 410M to 3.69 at 12B. The negative exponent matches the Cerebras-GPT pattern (Figure~\ref{fig:scaling_related}), confirming cross-family generalization of improved semantic filtering at scale; the lower $R^2$ compared to Counterfactual ($R^2 = 0.998$) reflects the weaker scaling signal in Related contexts. (Right) Irrelevant: distractor entrainment increases from 2.09 at 410M to 2.72 at 12B. The positive exponent is consistent with the Cerebras-GPT findings (Figure~\ref{fig:scaling_irrelevant}), demonstrating that mechanical copying generalizes across model families; the higher $R^2$ in Pythia suggests more regular scaling behavior for this condition. Shaded regions show 95\% confidence intervals.}
\label{fig:pythia_scaling_rel_irr}
\end{figure*}

Figures~\ref{fig:pythia_scaling_combined}--\ref{fig:pythia_scaling_irrelevant} present the Pythia scaling analyses. The combined plot (Figure~\ref{fig:pythia_scaling_combined}) directly parallels the main text Figure~2 for Cerebras-GPT, enabling visual comparison of the sign split across model families.

%==============================================================================
\section{Convergence Analysis}\label{app:convergence}
%==============================================================================

To understand the relative dynamics between gold and distractor tokens, we plot both metrics jointly across model scale. The gap between curves indicates the model's net preference for correct answers over distractors---a direct measure of functional accuracy under contextual influence. These analyses complement the Related and Random convergence plots shown in the main text (Figure~3) by presenting the Irrelevant and Counterfactual conditions.

The key insight from convergence analysis is that context type determines whether scaling \textit{helps} or \textit{hurts} model discrimination. For semantic contexts, the gold-distractor gap narrows favorably (convergent behavior): larger models simultaneously suppress distractors and boost correct answers. For non-semantic contexts, the gap widens unfavorably (divergent behavior): larger models increasingly favor distractors over gold tokens. This convergent-divergent split reinforces the dual-mechanism interpretation from the main text.

%------------------------------------------------------------------------------
\subsection{Cerebras-GPT (111M--13B)}
%------------------------------------------------------------------------------

\begin{figure*}[t]
\centering
\begin{subfigure}{0.48\textwidth}
\centering
\includegraphics[width=\linewidth]{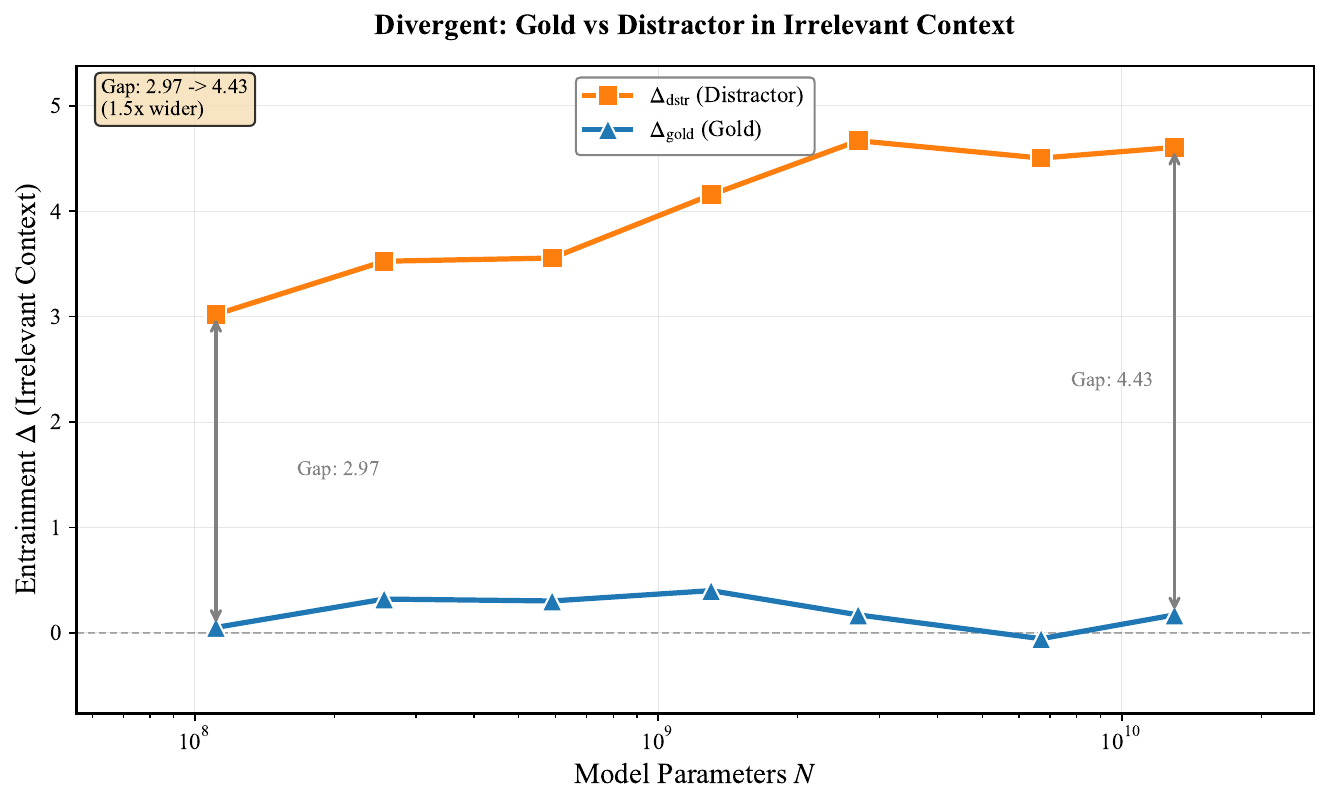}
\caption{Irrelevant context: divergent behavior.}
\label{fig:conv_irrelevant}
\end{subfigure}
\hfill
\begin{subfigure}{0.48\textwidth}
\centering
\includegraphics[width=\linewidth]{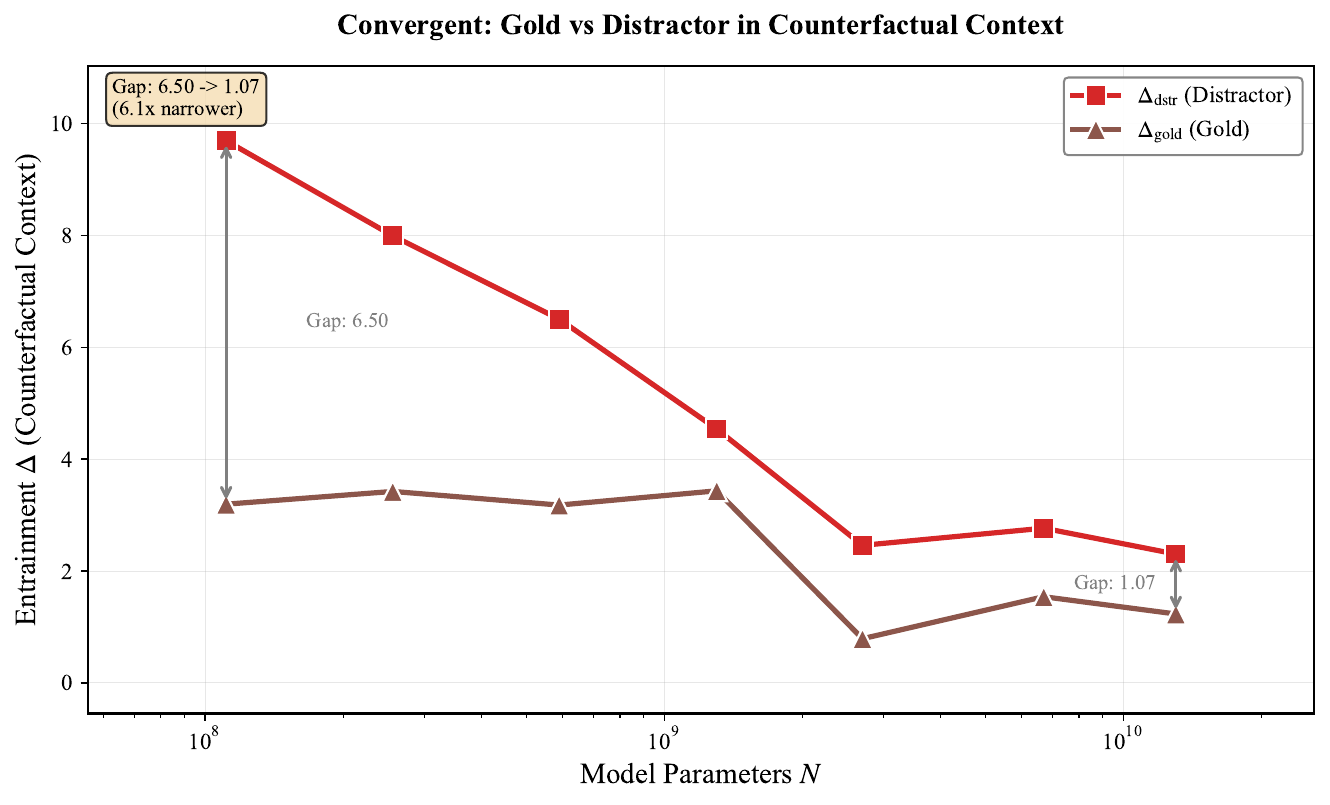}
\caption{Counterfactual context: convergent behavior.}
\label{fig:conv_counterfactual}
\end{subfigure}
\caption{\textbf{Cerebras-GPT: Convergence vs.\ divergence for Irrelevant and Counterfactual contexts.} Joint scaling of gold entrainment ($\Delta_{\mathrm{gold}}$, blue) and distractor entrainment ($\Delta_{\mathrm{dstr}}$, red). (Left) Irrelevant: gold entrainment is essentially flat ($b_{\mathrm{gold}} \approx 0$, ranging from 0.05 to 0.17) while distractor entrainment increases ($b_{\mathrm{dstr}} = +0.09$, from 3.02 to 4.61); the gap widens from 2.97 at 111M to 4.44 at 13B (1.5$\times$ increase). This \textbf{divergent} pattern indicates that models become relatively more distracted by irrelevant context at scale---because Irrelevant contexts lack semantic content, the filtering mechanism does not engage, leaving only mechanical copying to scale upward. (Right) Counterfactual: distractor entrainment drops sharply ($b_{\mathrm{dstr}} = -0.33$, from 9.69 to 2.30) while gold entrainment remains relatively stable (0.79 to 3.44); the gap narrows from 6.50 at 111M to 1.07 at 13B (6.1$\times$ reduction). This \textbf{convergent} pattern demonstrates effective misinformation resistance at scale: larger models not only suppress false claims but maintain or boost the correct answer.}
\label{fig:conv_irrelevant_counterfactual}
\end{figure*}

%------------------------------------------------------------------------------
\subsection{Pythia (410M--12B)}
%------------------------------------------------------------------------------

\begin{figure*}[t]
\centering
\includegraphics[width=\textwidth]{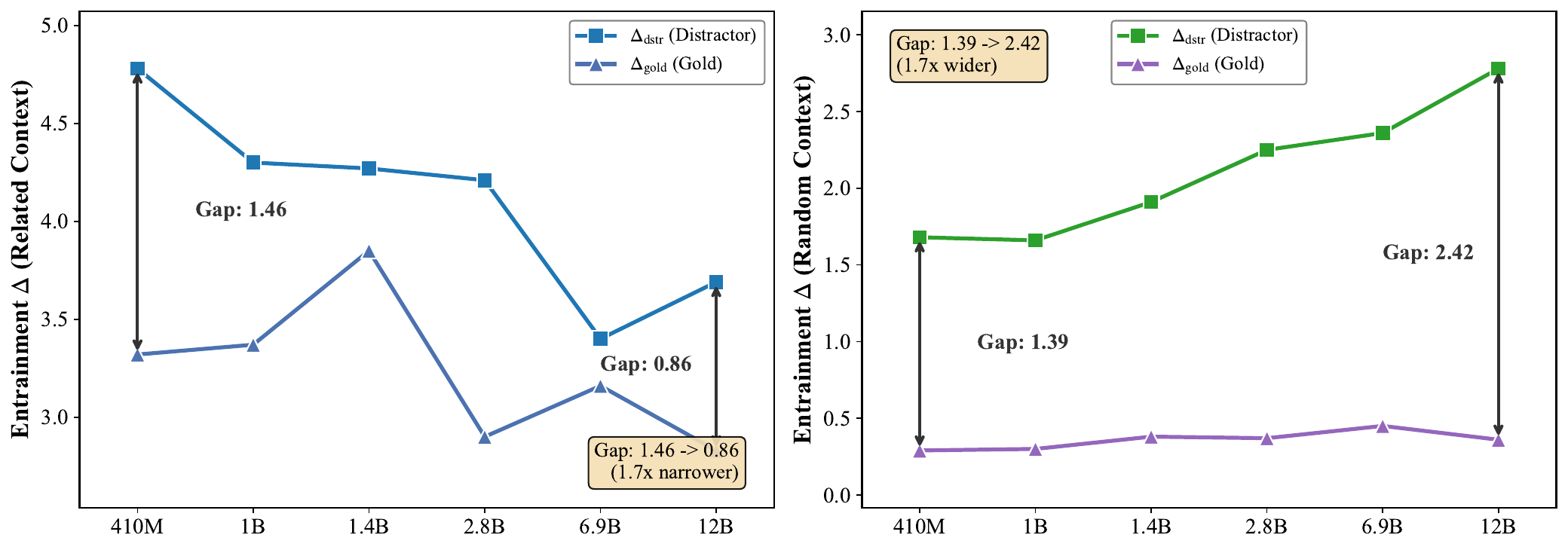}
\caption{\textbf{Pythia: Convergence vs. Divergence Across Context Types.} Combined visualization showing opposite convergence behaviors for semantic vs. non-semantic contexts. (Left) Related context shows \textbf{convergent} behavior: the gold-distractor gap narrows with scale as distractor entrainment decreases ($b_{\mathrm{dstr}} = -0.09$) while gold entrainment remains stable. The gap reduces from 2.41 at 410M to 0.74 at 12B---a 3.3$\times$ improvement in discrimination. (Right) Random context shows \textbf{divergent} behavior: the gap widens with scale as distractor entrainment increases ($b_{\mathrm{dstr}} = +0.16$) while gold entrainment stays flat. The gap increases from 1.44 at 410M to 2.13 at 12B---a 1.5$\times$ degradation. This pattern exactly replicates the Cerebras-GPT findings (Figure~3 in main text), confirming that the convergent-divergent split is a general property of Transformer scaling, not a family-specific artifact. The replication across independently trained model families provides strong evidence for the dual-mechanism interpretation. Figure~\ref{fig:pythia_exp_conv_irr}(b) additionally presents the Pythia Irrelevant divergence plot in isolation.}
\label{fig:pythia_convergence_combined}
\end{figure*}

%==============================================================================
\section{Additional Visualizations}\label{app:additional}
%==============================================================================

This section provides supplementary visualizations that support the main findings through alternative representations. Log-log plots verify the power-law assumption underlying our scaling analysis, while heatmaps provide an intuitive overview of entrainment patterns across all conditions and model sizes.

%------------------------------------------------------------------------------
\subsection{Cerebras-GPT (111M--13B)}
%------------------------------------------------------------------------------

\begin{figure*}[t]
\centering
\begin{subfigure}{0.48\textwidth}
\centering
\includegraphics[width=\linewidth]{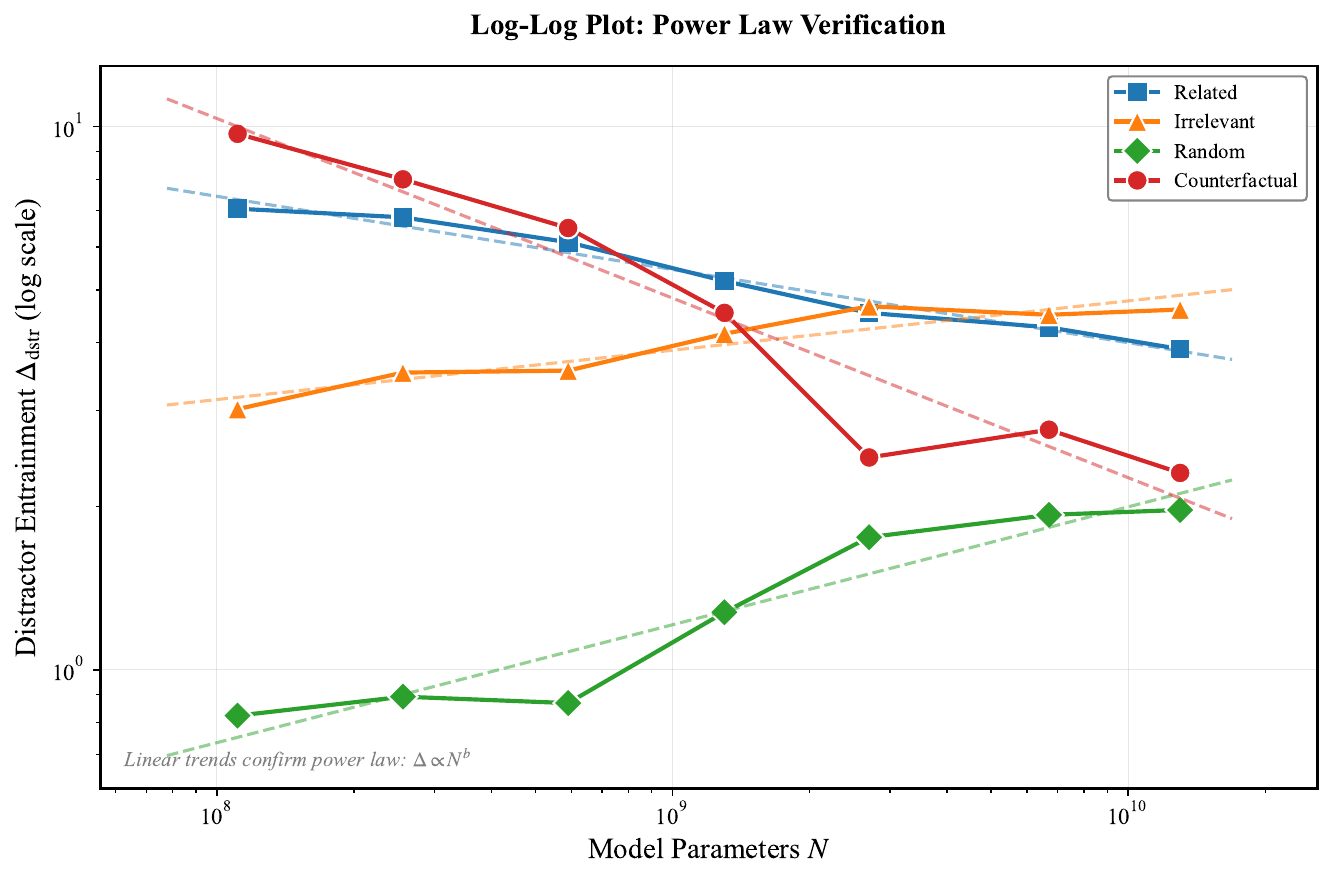}
\caption{Log-log verification of power-law scaling.}
\label{fig:loglog}
\end{subfigure}
\hfill
\begin{subfigure}{0.48\textwidth}
\centering
\includegraphics[width=\linewidth]{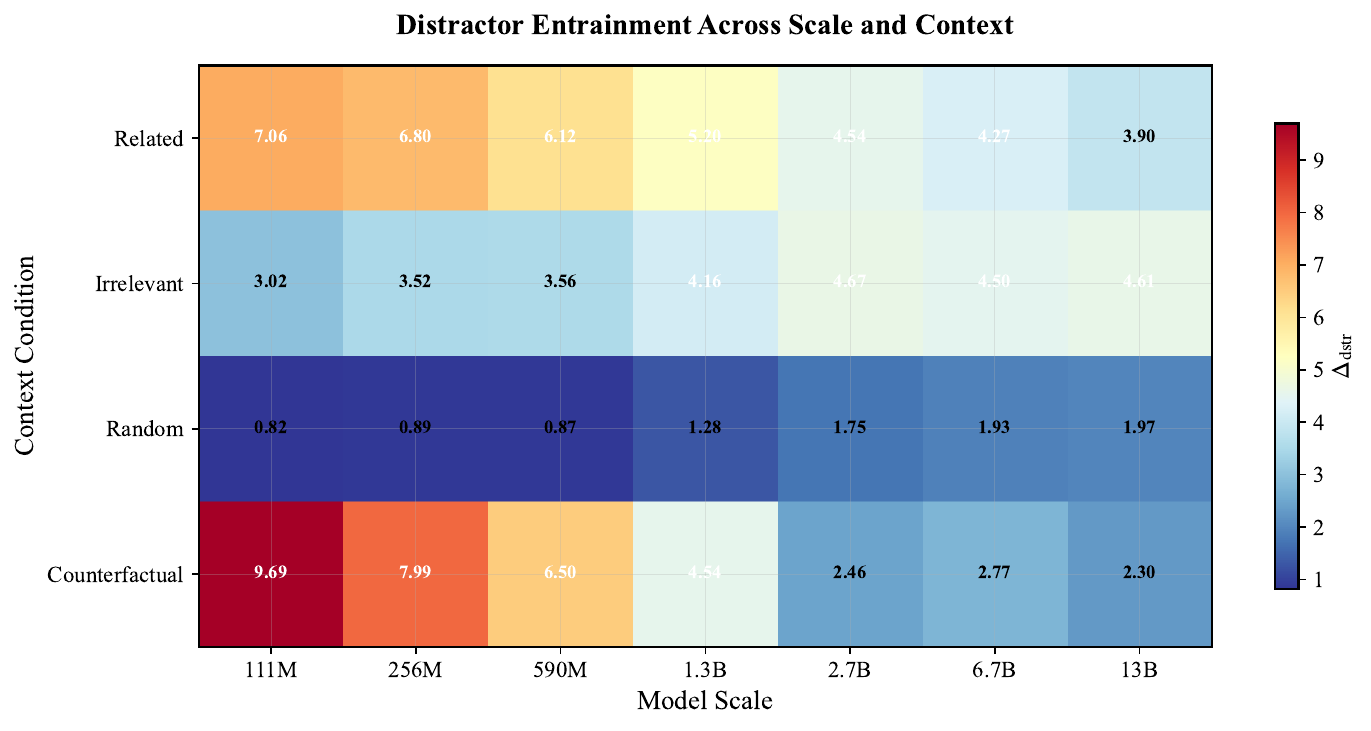}
\caption{Distractor entrainment heatmap.}
\label{fig:heatmap}
\end{subfigure}
\caption{\textbf{Cerebras-GPT: Log-log verification and entrainment heatmap.} (Left) Distractor entrainment ($\Delta_{\mathrm{dstr}}$) plotted against model size ($N$) in log-log space for all four context conditions. Power-law relationships $\Delta_{\mathrm{dstr}} \propto N^b$ appear as straight lines with slope $b$. Semantic contexts (Counterfactual, Related) show negative slopes (downward trends), while non-semantic contexts (Irrelevant, Random) show positive slopes (upward trends). The linearity across nearly two orders of magnitude in model size (111M--13B) confirms that power laws accurately describe entrainment scaling, and the parallel structure of lines within each group suggests a common underlying mechanism with condition-specific magnitudes. (Right) Matrix visualization of $\Delta_{\mathrm{dstr}}$ across all combinations of context condition (rows) and model size (columns, 111M--13B). Color intensity indicates entrainment magnitude. The heatmap reveals two distinct gradients: semantic contexts decrease from left to right, while non-semantic contexts increase. The Counterfactual row shows the steepest gradient (strongest negative scaling), while the Random row shows clear brightening with scale (strongest positive scaling).}
\label{fig:loglog_heatmap}
\end{figure*}

\paragraph{Log-Log Verification and Entrainment Heatmap.}
Figure~\ref{fig:loglog_heatmap} presents the Cerebras-GPT data in log-log space alongside a matrix view of distractor entrainment values across all model sizes and context conditions. Together they verify the power-law assumption and provide an intuitive visualization of the divergent scaling patterns.

%------------------------------------------------------------------------------
\subsection{Pythia (410M--12B)}
%------------------------------------------------------------------------------

\begin{figure*}[t]
\centering
\begin{subfigure}{0.48\textwidth}
\centering
\includegraphics[width=\linewidth]{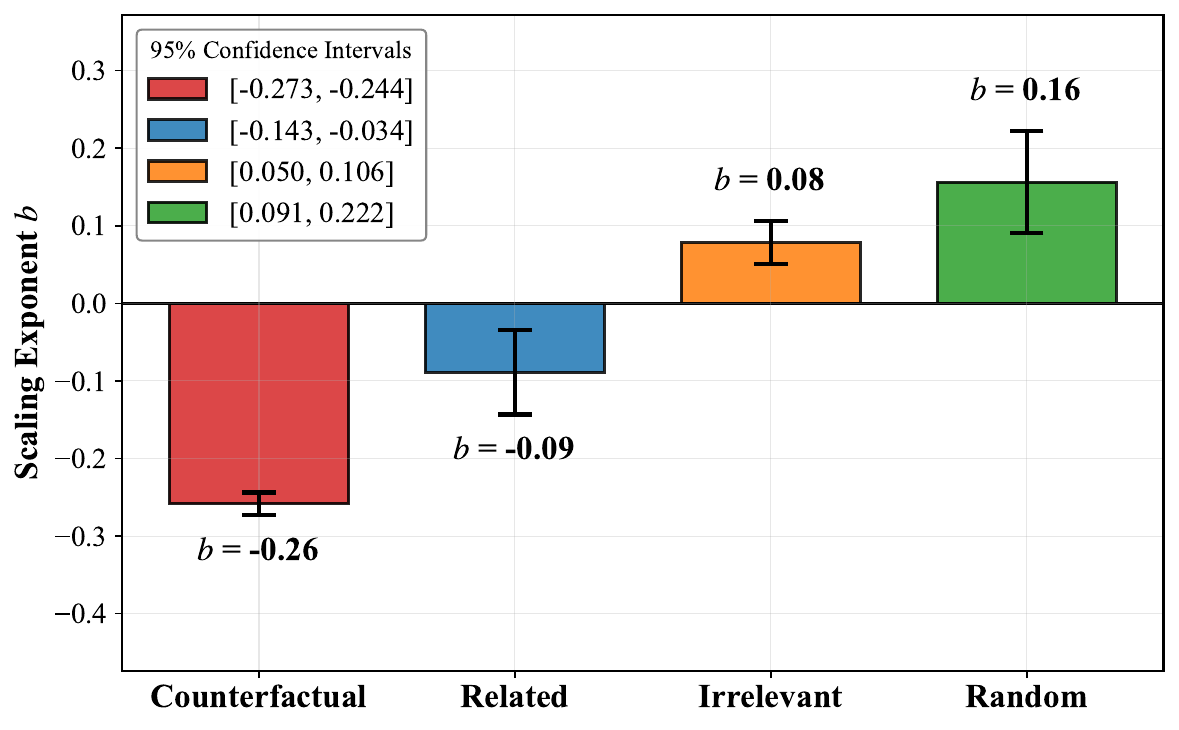}
\caption{Scaling exponents with 95\% confidence intervals.}
\label{fig:pythia_exponents}
\end{subfigure}
\hfill
\begin{subfigure}{0.48\textwidth}
\centering
\includegraphics[width=\linewidth]{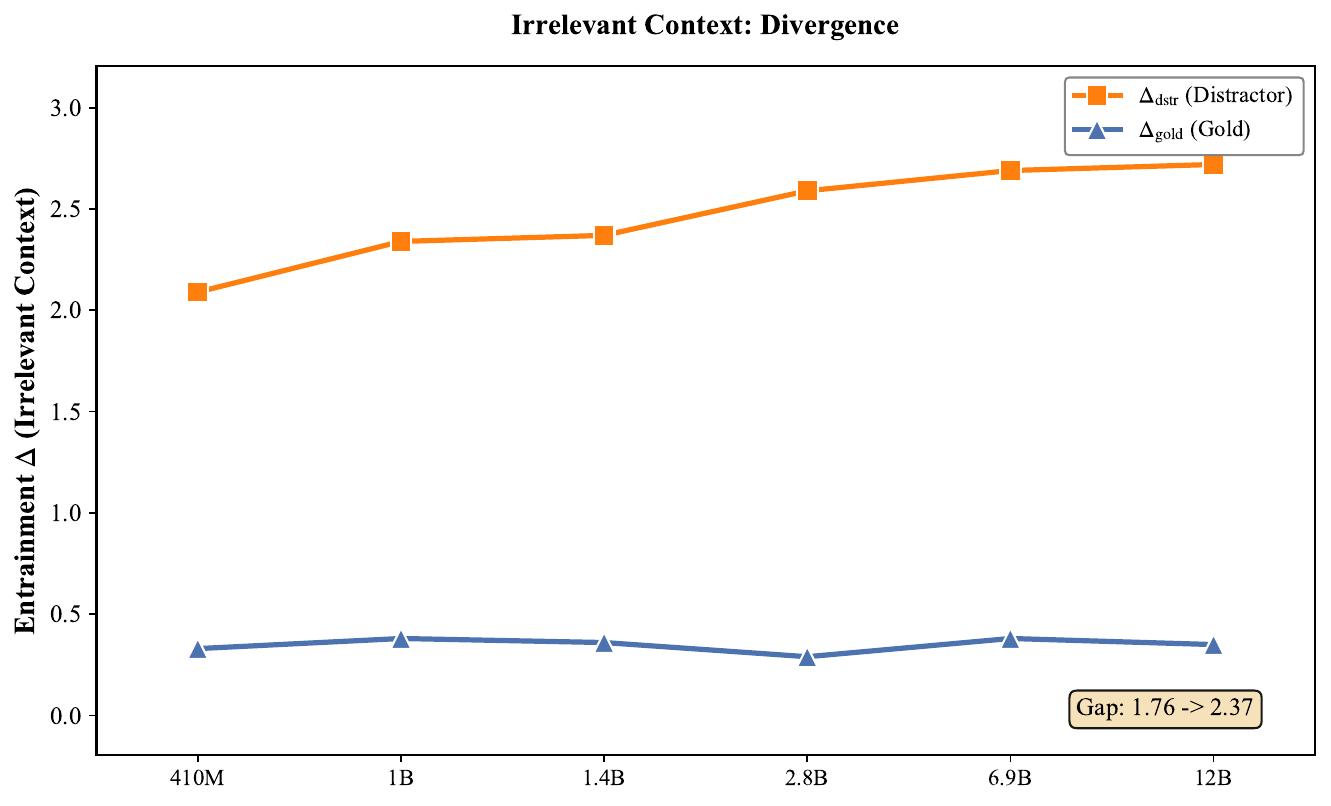}
\caption{Irrelevant context: divergent behavior.}
\label{fig:pythia_conv_irrelevant}
\end{subfigure}
\caption{\textbf{Pythia: Scaling exponents and Irrelevant-context divergence.} (Left) Estimated power-law exponents ($b$) for distractor entrainment ($\Delta_{\mathrm{dstr}}$) across the four context conditions, with 95\% confidence intervals shown as error bars. Semantic contexts (Counterfactual: $b = -0.26$; Related: $b = -0.09$) show negative exponents, while non-semantic contexts (Random: $b = +0.16$; Irrelevant: $b = +0.08$) show positive exponents. The confidence intervals for semantic and non-semantic groups do not overlap, establishing statistical separation. This pattern exactly replicates the Cerebras-GPT findings (Figure~1 in main text), with the same ordering of exponent magnitudes: $|b_{\mathrm{CF}}| > |b_{\mathrm{Rel}}|$ and $|b_{\mathrm{Rnd}}| > |b_{\mathrm{Irr}}|$. (Right) Joint scaling of gold entrainment ($\Delta_{\mathrm{gold}}$, blue) and distractor entrainment ($\Delta_{\mathrm{dstr}}$, red) for the Irrelevant context condition. Similar to Cerebras-GPT (Figure~\ref{fig:conv_irrelevant}), gold entrainment remains essentially flat across scale (0.29 to 0.38) while distractor entrainment increases ($b_{\mathrm{dstr}} = +0.08$, from 2.09 to 2.72); the gap widens from 1.89 at 410M to 2.22 at 12B. This \textbf{divergent} pattern replicates across model families, confirming that the mechanical copying mechanism scales independently of semantic content.}
\label{fig:pythia_exp_conv_irr}
\end{figure*}

\paragraph{Scaling Exponents and Irrelevant Divergence.}
Figure~\ref{fig:pythia_exp_conv_irr} shows the Pythia scaling exponents with 95\% confidence intervals alongside the Irrelevant-context convergence plot, enabling direct comparison with the Cerebras-GPT analysis presented in the main text.

\begin{figure*}[t]
\centering
\begin{subfigure}{0.48\textwidth}
\centering
\includegraphics[width=\linewidth]{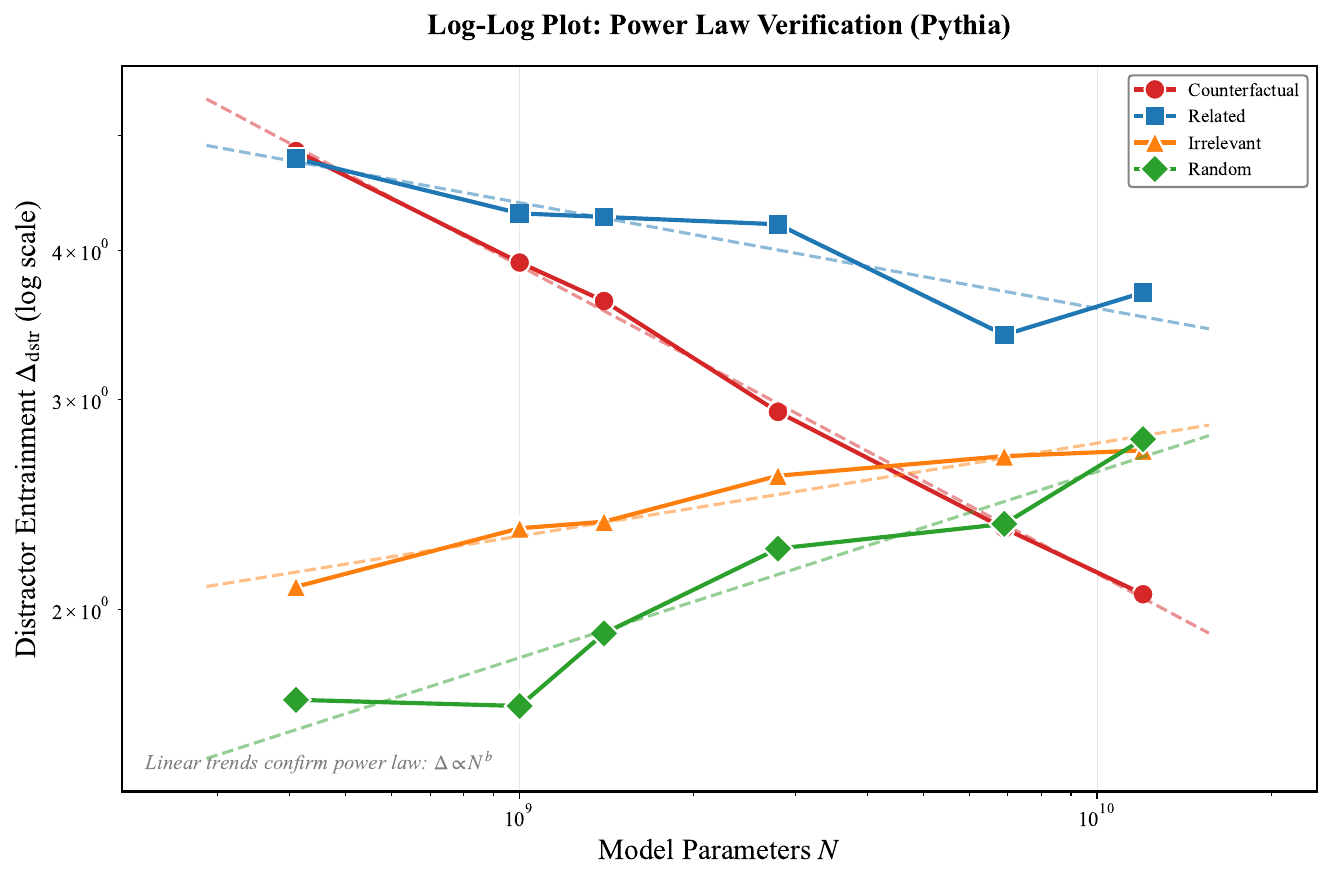}
\caption{Log-log verification of power-law scaling.}
\label{fig:pythia_loglog}
\end{subfigure}
\hfill
\begin{subfigure}{0.48\textwidth}
\centering
\includegraphics[width=\linewidth]{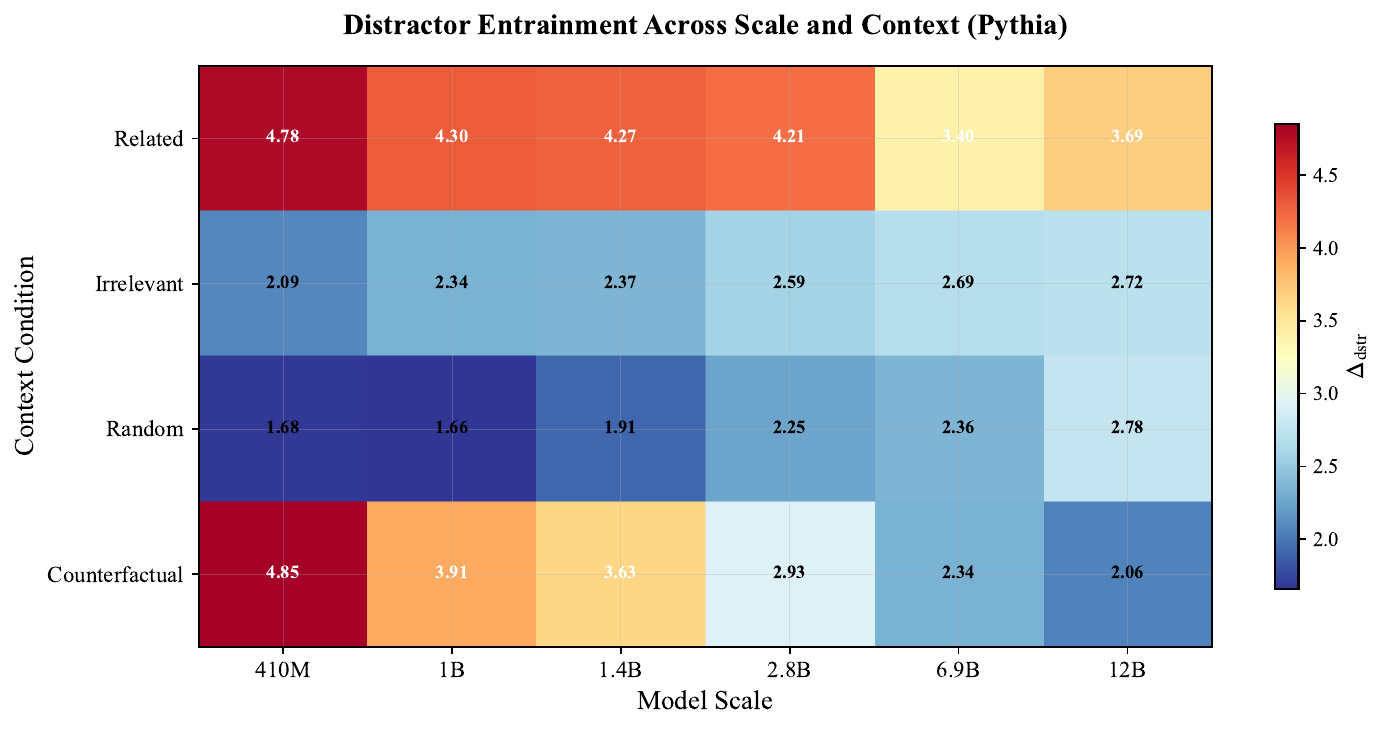}
\caption{Distractor entrainment heatmap.}
\label{fig:pythia_heatmap}
\end{subfigure}
\caption{\textbf{Pythia: Log-log verification and entrainment heatmap.} (Left) Distractor entrainment ($\Delta_{\mathrm{dstr}}$) plotted against model size ($N$) in log-log space for all four context conditions. As with Cerebras-GPT (Figure~\ref{fig:loglog}), power-law relationships appear as straight lines. The Counterfactual condition shows particularly tight linearity ($R^2 = 0.998$), providing strong evidence that misinformation resistance scales as a precise power law. The semantic/non-semantic split in slopes replicates across model families, confirming that $\Delta_{\mathrm{dstr}} \propto N^b$ with opposite-signed $b$ is a general scaling pattern. (Right) Matrix visualization of $\Delta_{\mathrm{dstr}}$ across all combinations of context condition (rows) and model size (columns, 410M--12B). The same divergent gradient pattern observed in Cerebras-GPT (Figure~\ref{fig:heatmap}) emerges here: semantic contexts (Counterfactual, Related) darken from left to right (decreasing entrainment), while non-semantic contexts (Irrelevant, Random) brighten (increasing entrainment). The overall lower intensity compared to Cerebras-GPT reflects Pythia's generally smaller entrainment magnitudes, though the directional trends are identical.}
\label{fig:pythia_loglog_heatmap}
\end{figure*}

\paragraph{Log-Log Verification and Entrainment Heatmap.}
Figure~\ref{fig:pythia_loglog_heatmap} presents the Pythia log-log verification alongside the entrainment heatmap, paralleling the corresponding Cerebras-GPT visualizations.

\end{document}